\documentclass[a4paper,fleqn]{cas-sc}

\usepackage[authoryear]{natbib}
\usepackage{hyperref}
\usepackage{amsmath,amssymb,amsfonts,bm,mathtools}
\usepackage{chngcntr}
\usepackage{booktabs}
\usepackage{microtype}
\usepackage{algorithm}
\usepackage{algpseudocode}
\usepackage{tikz}
\usepackage{array}
\usepackage{graphicx}
\usepackage{tabularx}
\usepackage{threeparttable}
\usepackage{makecell}
\usepackage{multirow}
\usepackage{pifont}
\usepackage{adjustbox}
\usepackage{xcolor}
\usepackage{placeins}
\usepackage{capt-of}
\usepackage{needspace}

\allowdisplaybreaks
\hypersetup{colorlinks=false,pdfborder={0 0 0},hypertexnames=false}

\newcommand{\V}{\mathcal{V}}
\newcommand{\B}{\mathcal{B}}
\newcommand{\Tbatch}{\mathcal{T}_{\mathcal{B}}}
\newcommand{\I}{\mathcal{I}}
\newcommand{\R}{\mathcal{R}}
\newcommand{\C}{\mathcal{C}}
\newcommand{\Sset}{\mathcal{S}}
\newcommand{\Pset}{\mathcal{P}}

\newcommand{\TopD}{\operatorname{TopD}}
\newcommand{\KL}{\operatorname{KL}}
\newcommand{\clip}{\operatorname{clip}}
\newcommand{\round}{\operatorname{round}}
\newcommand{\softmax}{\operatorname{softmax}}

\definecolor{alrared}{RGB}{190,35,35}
\definecolor{alragray}{RGB}{100,100,100}
\definecolor{alragreen}{RGB}{35,135,65}

\newcommand{\cmark}{\textcolor{alrared}{\ding{51}}}
\newcommand{\xmark}{\textcolor{alragray}{\ding{55}}}

\newcommand{\NA}{--}

\newcommand{\ALRA}{\textsc{ALRA}}
\newcommand{\ALRAR}{\textsc{ALRA}-R}
\newcommand{\ALRAU}{\textsc{ALRA}-U}
\newcommand{\ALRAG}{\textsc{ALRA}-G}

\newcommand{\ALRAN}{\textsc{ALRA}-N}
\newcommand{\ALRANG}{\textsc{ALRA}-NG}

\newcommand{\PRAU}{\textsc{PRA-U}}
\newcommand{\SWPRA}{\textsc{SWPRA}}

\newcommand{\FixedBudget}{Fixed-Budget}

\newcommand{\pubyear}[1]{\textcolor{alragray}{#1}}
\newcommand{\basecite}[1]{\textcolor{alragray}{\citep{#1}}}
\newcommand{\tabbest}[1]{\textbf{#1}}
\newcommand{\tabsecond}[1]{\underline{#1}}
\newcommand{\tabavgbest}[1]{\textcolor{alrared}{\textbf{#1}}}
\newcommand{\tabdeltabest}[1]{\textcolor{alrared}{\textbf{+#1}}}

\begin{document}
\let\WriteBookmarks\relax
\raggedbottom

\shorttitle{Adaptive Local Relational Alignment}

\shortauthors{Trung et~al.}

\title[mode=title]{ALRA: Adaptive Local Relational Alignment for
Logit-Based Pre-training Distillation of Autoregressive Language Models}


\author[1]{Quang Hoang Trung}
\fnmark[1]
\cormark[1]
\ead{trung.quang@vj-tech.jp}

\author[2,1]{Quang Huu Hieu}
\fnmark[1]
\ead{hieuquang@aj-tech.jp}

\author[1]{Nguyen Van Hoang Phuc}
\ead{phucnvh2310@gmail.com}

\author[1,3,4]{Vo Nguyen Le Duy}
\ead{duyvnl@vj-tech.jp}


\affiliation[1]{
    organization={VJ Technologies},
    city={Da Nang},
    country={Vietnam}
}

\affiliation[2]{
    organization={AJ Technologies},
    city={Nagoya},
    country={Japan}
}

\affiliation[3]{
    organization={Vietnam National University},
    city={Ho Chi Minh City},
    country={Vietnam}
}

\affiliation[4]{
    organization={University of Information Technology},
    city={Ho Chi Minh City},
    country={Vietnam}
}


\cortext[1]{Corresponding author.}

\fntext[1]{These authors contributed equally and are co-first authors.}

\begin{abstract}
Logit-based knowledge distillation for autoregressive language models usually
aligns teacher and student next-token distributions over the entire vocabulary.
However, this global objective overlooks relative preferences among likely
token alternatives. Existing local approaches often select candidate tokens
from either the teacher or the student alone. Teacher-only selection can miss
tokens that the student considers likely, while student-only selection can
rely on an inaccurate ranking early in training. We propose Adaptive Local
Relational Alignment (ALRA), a position-specific framework combining student
proposals with teacher guidance. At each valid prediction position, the
student proposes likely tokens, while the teacher's most probable token is
included as an anchor. ALRA adjusts the number of selected tokens according to
how broadly the teacher distributes probability within this candidate set
relative to the current batch. Adaptive Local Divergence retains the
mass-matching term and separately matches the relative token distributions
within the selected and remaining vocabulary regions. Unlike the exact
full-vocabulary decomposition, it replaces the teacher-mass coefficients of
the two conditional terms with unit coefficients, preventing either term from
being downweighted solely because its region has low teacher probability.
Student-Weighted Pairwise Relational Alignment emphasizes high-probability
token pairs with small student probability gaps and gives less weight to
unlikely or clearly separated pairs. Experiments on The Pile with randomly
initialized 200M- and 500M-parameter students across nine zero-shot benchmarks
yield average accuracies of 36.62\% and 37.40\%. ALRA exceeds the strongest
competing distillation baseline by 0.94 and 0.83 percentage points and
improves over pre-training without distillation by 2.31 and 2.91 points,
respectively.
\end{abstract}


\maketitle

\section{Introduction}
\label{sec:introduction}

Large language models (LLMs) have achieved strong performance across a broad
range of natural language processing tasks, but their increasing scale also
introduces substantial computational and memory requirements. Knowledge distillation (KD) provides an established
approach for transferring knowledge from a larger teacher model to a smaller
student model, allowing the student to learn from information beyond the
observed target token \citep{hinton2015distilling}. For autoregressive language
models, logits-based distillation typically minimizes the forward
Kullback--Leibler (KL) divergence between teacher and student next-token
distributions over the entire vocabulary. Although this
formulation provides comprehensive distribution-level supervision, it treats
all vocabulary tokens within a single global objective, despite the fact that
different prediction contexts can exhibit substantially different levels of
uncertainty. In particular, some contexts produce highly concentrated teacher
distributions, whereas others assign meaningful probability to several
competing alternatives. This motivates more selective and adaptive
distillation strategies that can allocate supervision according to the
characteristics of each prediction position.

\begin{figure}[pos=t]
    \centering
    \makebox[\textwidth][c]{%
        \includegraphics[
            width=0.97\textwidth,
            trim=0 0 17mm 0,
            clip
        ]{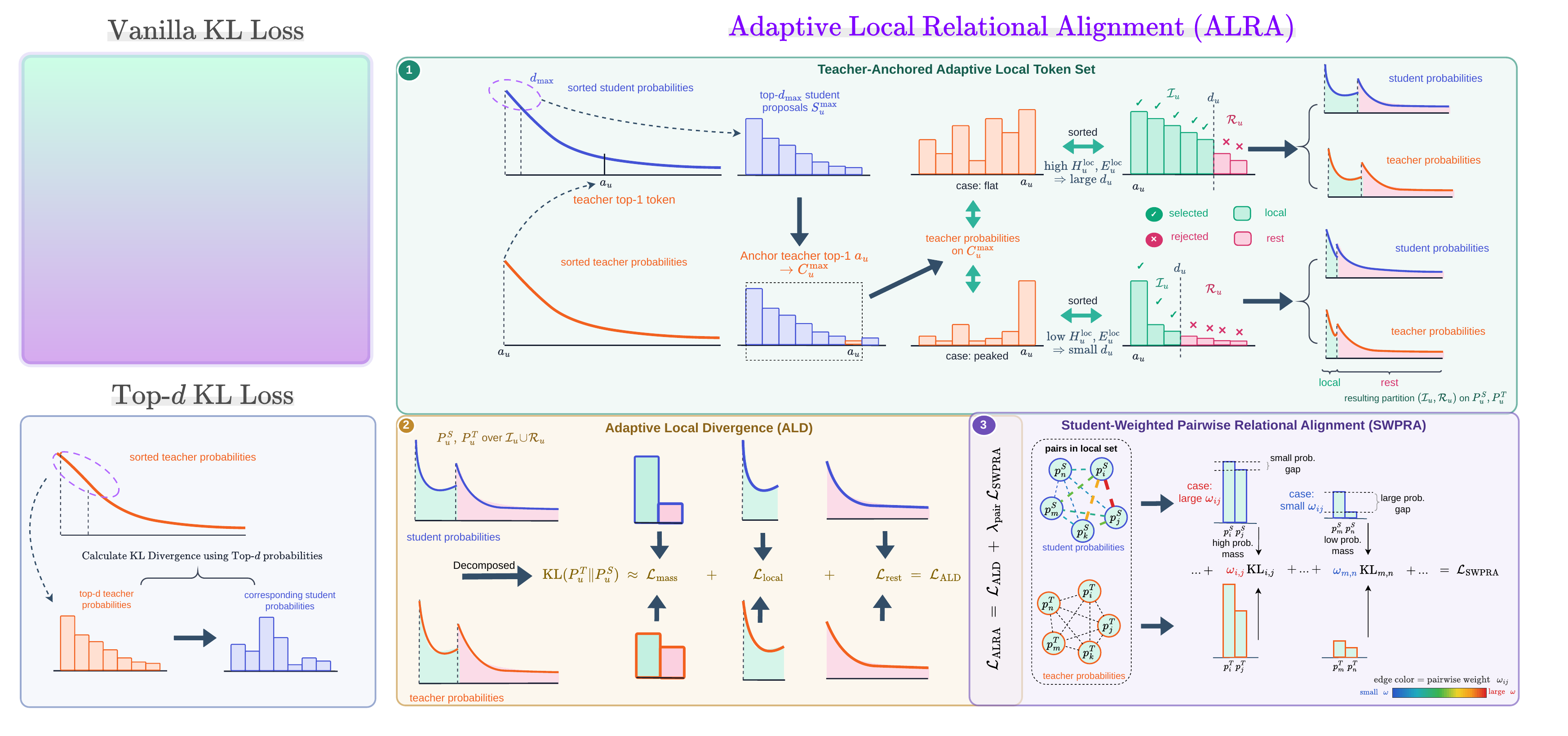}
    }
    \caption{
    Overview of Adaptive Local Relational Alignment (ALRA) and its comparison
    with Vanilla and Top-\(d\) KL Loss. Vanilla KL Loss aligns the complete
    teacher and student next-token distributions over the full vocabulary,
    whereas Top-\(d\) KL Loss selects the \(d\) highest-probability vocabulary
    tokens under the teacher distribution at each prediction position,
    retrieves the student probabilities at the corresponding vocabulary
    indices, renormalizes both restricted distributions over the selected
    support, and computes their KL divergence. For each valid next-token
    prediction position \(u\in\Tbatch\), ALRA first selects the \(d_{\max}\)
    vocabulary tokens with the largest probabilities under the student's next-token distribution \(P_u^S\) and
    ensures inclusion of the teacher top-1 token to form a teacher-anchored
    candidate set. It then computes the effective support size of the teacher
    distribution renormalized within this set and compares it with the
    batch-average effective support size to determine the position-specific
    local-set size \(d_u\). The final local set \(\I_u\) contains the \(d_u\)
    highest-probability tokens under the teacher distribution within the
    anchored candidate set, and \(\R_u=\V\setminus\I_u\) is its
    full-vocabulary complement. Adaptive Local Divergence (ALD) uses the mass,
    local-conditional, and rest-conditional structures identified by a
    local--rest decomposition of the full-vocabulary forward KL as separate
    supervision components, while assigning unit coefficients to the two
    conditional components. Within the adaptive local set \(\I_u\),
    Student-Weighted Pairwise Relational Alignment (SWPRA) assigns larger
    weights to token pairs with high total probability mass under the
    student's original full-vocabulary distribution and small student
    probability gaps, thereby emphasizing pairwise supervision for plausible
    tokens that the student currently separates only weakly. Conversely,
    low-mass pairs and pairs with large student probability gaps receive less
    emphasis. The final objective combines ALD and SWPRA and updates only the
    student model.
    }
    \label{fig:alra_pipeline}
\end{figure}

A central challenge in selective distillation is how to identify the tokens
that should receive detailed supervision while avoiding excessive dependence
on either the teacher or the student. Existing restricted-logit approaches
include teacher-based truncation \citep{peng2025pre} and student-based local
selection \citep{xu2025local}. Teacher-only selection may overlook tokens that
the student currently assigns high probability, whereas student-only selection
can be unreliable when the student's ranking is still inaccurate during early
training. Moreover, a fixed local-set size keeps the number of selected tokens
the same across prediction positions, although some contexts require only a few
alternatives while others involve a broader set of high-probability tokens.
Recent studies have investigated position-selective, relational,
difficulty-aware, and tail-aware distillation
\citep{tavor2026rethinking,xu2025local,he2025kd,dasguptadon}.
Pairwise relational supervision captures relative preferences among competing
tokens \citep{xu2025local}, but applying such relations to autoregressive
language modeling requires position-specific handling of a large and
context-dependent vocabulary. These observations point to the need for a
unified distillation objective that can adapt the supervised local region to
each prediction context while preserving information from the remaining
vocabulary and explicitly modeling relationships among high-probability
alternatives.

To address this challenge, we propose Adaptive Local Relational Alignment (ALRA), a position-specific logit-distillation framework that combines adaptive token selection, region-aware divergence, and student-dependent pairwise relational supervision. The main contributions of this paper are summarized as follows:
\begin{itemize}

\item We introduce a teacher-anchored adaptive local token selection mechanism for autoregressive language models. At each prediction position, the student proposes a
candidate set, while the teacher top-1 token is included as an anchor.
The teacher effective support within this candidate set, relative to
its current batch average, determines the local budget \(d_u\). The final local set is then formed by selecting the highest-probability tokens under the teacher distribution within the anchored candidate set. This design
keeps the candidate set responsive to the student's current predictions
while providing teacher guidance and allowing the local-set size to
vary across prediction positions.

\item We introduce two complementary objectives for the resulting
local--rest partition. Adaptive Local Divergence (ALD) starts from the
exact local--rest decomposition of forward KL, retains the mass-matching
term, and assigns unit coefficients to the local- and rest-conditional
divergences so that neither conditional term is directly scaled down by
its teacher region mass. Student-Weighted Pairwise Relational Alignment
(SWPRA) further aligns relative preferences between local token pairs,
giving more emphasis to high-probability alternatives that the student
currently separates only weakly.

\item We evaluate ALRA in from-scratch pre-training distillation from a
frozen Qwen1.5-1.8B teacher to randomly initialized 200M- and
500M-parameter students on The Pile. Across nine zero-shot benchmarks,
ALRA achieves average accuracies of 36.62\% and 37.40\%, exceeding the
strongest compared distillation baseline by 0.94 and 0.83 percentage
points, respectively. Controlled studies further compare the complete
ALRA configuration with fixed local budgets and separately examine
candidate-set anchoring and pairwise weighting.

\end{itemize}

\section{Related Work}
\label{sec:related}

\paragraph{Distillation for large language models.}
Knowledge distillation (KD) \citep{hinton2015distilling} is a common approach for compressing large language models (LLMs) into smaller students, thereby reducing inference cost and memory footprint while seeking to preserve predictive performance \citep{yang2025survey}. Recent surveys characterize the LLM distillation landscape from complementary perspectives. \citet{yang2025survey} classify LLM distillation methods into white-box and black-box KD, further distinguishing logits-based and hint-based approaches within white-box KD, while black-box KD includes in-context learning (ICL), chain-of-thought (CoT), and instruction following, where only the teacher's outputs are accessible through an API rather than its internal logits or representations. \citet{fang2025knowledge} instead place knowledge distillation (KD) and dataset distillation (DD) within a unified framework that bridges \emph{model-centric and data-centric} perspectives, treating them as complementary paradigms. Their KD taxonomy covers rationale-based, uncertainty-aware, multi-teacher, dynamic/adaptive, and task-specific approaches, while their DD discussion considers optimization-based distillation and synthetic data generation alongside complementary data-selection strategies \citep{fang2025knowledge}. Within these taxonomies, our work falls on the model-centric side and focuses specifically on white-box, logits-based KD for causal LMs. We focus on logits-based rather than hint-based KD, the latter of which additionally supervises the student through intermediate representations. Representative encoder-based distillation methods transfer different forms of intermediate knowledge, including hidden-state and attention information, as exemplified by DistilBERT, TinyBERT, MiniLM, and MobileBERT \citep{sanh2019distilbert,jiao2020tinybert,wang2020minilm,sun2020mobilebert}. Direct pointwise hidden-state matching is architecture-sensitive because losses such as MSE or cosine similarity require dimensionally compatible representations or an explicit alignment mapping. TinyBERT, for example, introduces a learnable linear projection to align student and teacher hidden states when their hidden sizes differ \citep{jiao2020tinybert}. Conversely, TAD's larger-model experiments use students with the same hidden dimensionality as their teachers and directly apply an auxiliary cosine loss between teacher and student hidden states \citep{dasguptadon}. Because our students differ from the teacher in hidden size but share the same output vocabulary, we restrict the distillation signal to the output level, avoiding explicit alignment of intermediate representation spaces. Within this setting, we examine how structured, relational, tail-aware, and adaptive objectives affect the allocation of supervision across token positions and vocabulary entries while keeping the training pipeline otherwise fixed.

\paragraph{White-box logits-based KD for LLMs.}
A canonical white-box logits-based baseline matches the full teacher and student next-token distributions using the forward KL divergence \citep{hinton2015distilling,muralidharan2024compact}; we refer to this objective as Vanilla KD. Earlier NLP work also demonstrated output-level distillation across different architectures: \citet{tang2019distilling} transferred task-specific knowledge from BERT to a single-layer BiLSTM by minimizing the MSE between teacher and student logits. PD further studies pre-training distillation systematically across logits processing, loss selection, scaling law, and offline versus online teacher logits \citep{peng2025pre}. In the standard full-distribution forward-KL baseline, a single global divergence is applied to the complete teacher distribution at each token. Although the teacher's non-target probabilities contain information beyond the ground-truth token \citep{zhong2024revisiting}, this objective does not explicitly partition the output distribution into high- and low-probability regions or adapt the teaching strategy across token positions. These properties motivate structured, adaptive, and tail-aware extensions that retain the white-box, logit-level setting while controlling how teacher information is allocated across token positions and vocabulary regions.

\paragraph{Structured and relational KD.} 
Beyond pointwise matching, relational KD transfers structural information by preserving relationships within the teacher signal. RKD operates in representation space, preserving distance- and angle-based relations among training examples rather than matching individual representations independently \citep{park2019relational}. In pre-trained language model distillation, ReAugKD introduces a relationship loss that preserves semantic similarities among teacher and student training examples to support retrieval-augmented knowledge transfer \citep{zhang2023reaugkd}. At the output level, structured logit objectives instead model dependencies among classes. For image classification, RLD uses ground-truth labels to dynamically refine teacher logits, removing misleading teacher information while preserving class correlations \citep{sun2025knowledge}, whereas LDRLD recursively decouples and recombines the top-$d$ logits selected according to the student's ranking to capture fine-grained inter-class relations and adaptively emphasize critical category pairs \citep{xu2025local}. Taken together, these works illustrate relational structure at two levels: relationships among examples in representation space and dependencies among classes in the output distribution. Our work follows the latter direction in autoregressive LMs, modeling structured relationships among vocabulary entries while keeping the distillation signal at the output level.
\paragraph{Adaptive, selective, and tail-aware logit KD.} 
A complementary line of work studies where and how distillation supervision should be applied within logit-level KD. DA-KD dynamically adjusts the distillation dataset according to sample difficulty and introduces a bidirectional discrepancy loss to improve learning from difficult samples \citep{he2025kd}. Selective KD studies this question at finer granularity: SE-KD uses student entropy to select token positions for distillation, while SE-KD$_{3X}$ extends selection jointly across token positions, vocabulary classes, and training samples \citep{tavor2026rethinking}. In autoregressive LMs, ATKD decomposes token-level KL into target-oriented and diversity-oriented knowledge and uses a teacher-uncertainty coefficient to identify token difficulty, applying different teaching modes to easy- and hard-to-learn tokens \citep{zhong2024revisiting}. AdaKD further adapts the distillation process to each token's learning state by combining loss-driven adaptive token focusing with token-level inverse difficulty temperature scaling, both driven by a unified token-difficulty metric \citep{xie2026llm}. A related line focuses on how distillation supervision is distributed across the vocabulary. BiLD filters long-tail noise by retaining only the top-$k$ teacher and student logits and constructs bidirectional logit-difference terms to leverage their internal ranking information \citep{li2025bild}. TAD instead introduces a tail-aware divergence that decouples the teacher's top-$K$ probabilities from the remaining tail, increasing the contribution of lower-probability predictions \citep{dasguptadon}. Together, these methods adapt, select, or redistribute distillation supervision across samples, token positions, and vocabulary regions, motivating objectives that jointly control where and how teacher information is transferred.

\paragraph{Sequence-level, data-centric, and policy-based KD.} 
Beyond token-level distribution matching, other approaches transfer teacher knowledge through generated sequences, the pre-training data distribution, or student-generated trajectories. In neural machine translation, sequence-level KD (SeqKD) uses beam search to generate target sequences from the teacher and then trains the student with cross-entropy on the resulting teacher-generated dataset \citep{kim2016sequence}. In LLM pre-training, MiniPLM takes an offline, data-centric approach: its Difference Sampling strategy uses the discrepancy between a teacher and a small reference LM to refine the pre-training corpus, down-sampling easy and common instances, up-sampling hard and diverse instances, and filtering noisy or harmful examples, after which the student is trained from scratch with the standard next-token cross-entropy objective \citep{gu2024miniplm}. Policy-based approaches instead optimize distillation over student-generated outputs. MiniLLM minimizes the sequence-level reverse KL divergence between student and teacher distributions and derives a policy-gradient optimization procedure over student-generated sequences \citep{gu2024minillm}. GKD generalizes this on-policy perspective by allowing a mixture of fixed and student-generated sequences for distillation under teacher feedback, while supporting multiple divergences, including forward KL, reverse KL, and generalized Jensen--Shannon divergence, as well as integration with reinforcement-learning fine-tuning \citep{agarwal2024onpolicy}. More generally, at the sequence-distribution level, $f$-DISTILL formulates sequence-level KD as generalized $f$-divergence minimization, shows that SeqKD and related approaches can be viewed as approximations of variants within this framework, and derives a step-wise decomposition that reduces intractable sequence-level divergences to tractable word-level losses \citep{wen2023f}. These approaches modify the sequences, data distributions, or policies through which teacher knowledge is transferred, whereas our study focuses on the design of token-level white-box logit objectives under a common pre-training regime. We therefore treat them as complementary directions rather than direct baselines; in particular, data-centric, sequence-level, or policy-based strategies could be combined with our objective in future work.

\paragraph{Theoretical and empirical perspectives on distillation.} 
A parallel body of work examines why teacher distributions can improve student learning and which factors determine whether distillation is effective. In autoregressive language models, the teacher's distribution over non-target tokens conveys diversity-oriented information beyond the ground-truth token \citep{zhong2024revisiting}. From a statistical perspective, \citet{menon2021statistical} show that a Bayes teacher providing the true class probabilities can reduce the variance of the student's learning objective, and derive a bias--variance trade-off that characterizes how approximate teacher probability estimates affect student generalization. The teacher--student capacity gap is another important factor in distillation effectiveness: intermediate teacher-assistant models can mitigate large capacity gaps \citep{mirzadeh2020improved}, while \citet{zhang2025towards} find that, in language model distillation, the optimal teacher scale grows approximately linearly with student scale. At the same time, stronger imitation of the teacher need not imply better generalization: \citet{stanton2021does} show that increased teacher--student fidelity does not always improve student test performance and, in a self-distillation setting, enlarging the distillation set can increase fidelity while reducing test accuracy. These findings caution against treating teacher--student fidelity as a direct proxy for downstream gains and motivate our decomposition-based analysis of which components of the teacher distribution different distillation objectives emphasize.

\paragraph{Positioning of our approach.}
Taken together, prior work has explored several complementary design axes in
logit-level KD, including relational structure among output classes
\citep{sun2025knowledge,xu2025local}, token-adaptive supervision
\citep{zhong2024revisiting,xie2026llm}, multi-axis selection across token
positions, vocabulary classes, and training samples
\citep{tavor2026rethinking}, and explicit handling of low-probability or tail
logits, either by filtering them or by increasing their contribution to
distillation \citep{li2025bild,dasguptadon}. To our knowledge, existing
logit-level KD methods for autoregressive LMs do not jointly combine a
position-specific adaptive local vocabulary budget, explicit supervision of
both the selected local region and its full-vocabulary complement, and
pairwise relational weighting within the selected local set. ALRA targets
this intersection by (i) using an exact local--rest decomposition of the
forward KL to expose mass, local-conditional, and rest-conditional
supervision, (ii) constructing a teacher-anchored, position-specific adaptive
local set, and (iii) applying student-dependent pairwise relational weighting
within that set. We compare ALRA with representative logit-level distillation
baselines under a shared pre-training pipeline. Controlled studies
additionally compare the complete ALRA configuration with fixed local budgets
and separately examine candidate-set anchoring and pairwise weighting.
Accordingly, we interpret the main baseline comparison at the level of
complete objectives and reserve mechanism-specific conclusions for the
corresponding controlled analyses.

\section{Proposed Approach}
\label{sec:proposal}

We present Adaptive Local Relational Alignment (ALRA), a position-specific
logit-distillation framework for autoregressive language models. The overall
workflow is illustrated in Fig.~\ref{fig:alra_pipeline}. ALRA is motivated by
two complementary lines of prior work. First, token-level forward KL admits a
decomposition under a partition of the output space, exposing the probability
mass assigned to different regions and the conditional distributions within
those regions
\citep{zhao2022decoupled,zhong2024revisiting,dasguptadon}. Second,
pairwise logit relations can explicitly represent relative preferences
between locally competing tokens \citep{xu2025local}. A relational
construction designed for fixed-class classification cannot be applied
directly to language modeling, because each autoregressive prediction
position produces a context-dependent distribution over a large vocabulary
and may have a different level of local ambiguity.

ALRA makes two main contributions. First, it constructs an adaptive local
token set separately for every valid next-token prediction position. The
student proposes the \(d_{\max}\) highest-probability vocabulary tokens, the
teacher top-1 token is included as a teacher-preferred reference, and the
effective support size of the teacher distribution renormalized within this
anchored candidate set determines a position-specific local-set size relative
to the current forward batch. The final local set is selected by teacher
probability ranking within the anchored candidate set. Second, ALRA uses a
position-specific local--rest decomposition of the full-vocabulary forward KL
to identify mass, local-conditional, and rest-conditional supervision
structures. ALD then forms a distinct region-aware objective by retaining the
mass-matching term and assigning unit coefficients to the two conditional
terms, so their contributions are not directly attenuated by the teacher
probability masses of the corresponding regions. ALRA further augments this
objective with student-weighted pairwise relational alignment inside the
adaptive local token set. The pairwise weighting prioritizes high-mass local
alternatives that the student separates only weakly.

\subsection{Preliminaries}
\label{subsec:preliminaries}

Let \(\B\) denote the current forward batch, containing \(B\) tokenized
sequences. We define the set of valid next-token prediction positions as
\begin{equation}
    \Tbatch
    =
    \{(b,t):m_{b,t}=1\},
    \label{eq:valid_token_positions}
\end{equation}
where \(b\) indexes a sequence in the batch, \(t\) indexes a prediction
position within that sequence, and \(m_{b,t}\in\{0,1\}\) indicates whether
the prediction at position \((b,t)\) contributes to the training loss. Each
\(u=(b,t)\in\Tbatch\) therefore represents one valid next-token prediction
position.

As a special case, consider \(B\) unpadded sequences, each containing \(T\)
tokens, with no ignored labels. Under the standard one-token causal shift,
where the output at position \(t\) predicts the token at position \(t+1\),
each sequence contributes \(T-1\) valid prediction positions. Hence,
\[
    |\Tbatch|=B(T-1).
\]
ALRA computes its adaptive statistics over the valid positions in \(\Tbatch\)
within each forward pass.

Let \(\V\) denote the vocabulary, with size \(|\V|\). For each
\(u\in\Tbatch\), the teacher and student produce vocabulary-logit vectors
\(z_u^T,z_u^S\in\mathbb{R}^{|\V|}\). At distillation temperature \(\tau>0\),
their next-token distributions are
\begin{equation}
\begin{aligned}
    P_u^T
    &=
    \softmax(z_u^T/\tau),
    &
    p_{u,i}^T
    &=
    \frac{\exp(z_{u,i}^T/\tau)}
    {\sum_{j\in\V}\exp(z_{u,j}^T/\tau)},\\
    P_u^S
    &=
    \softmax(z_u^S/\tau),
    &
    p_{u,i}^S
    &=
    \frac{\exp(z_{u,i}^S/\tau)}
    {\sum_{j\in\V}\exp(z_{u,j}^S/\tau)},
    \qquad i\in\V.
\end{aligned}
\label{eq:distillation_distributions}
\end{equation}
Here, \(p_{u,i}^T\) and \(p_{u,i}^S\) denote the probabilities assigned by
the teacher and student, respectively, to vocabulary token \(i\) at prediction
position \(u\).

The conventional token-level distillation objective is the forward
Kullback--Leibler divergence between the teacher and student distributions
over the full vocabulary:
\begin{equation}
    \mathcal{L}_{\mathrm{KL}}(u)
    =
    \KL(P_u^T\|P_u^S)
    =
    \sum_{i\in\V}
    p_{u,i}^T
    \log
    \frac{p_{u,i}^T}{p_{u,i}^S}.
    \label{eq:standard_forward_kl}
\end{equation}

\subsection{Teacher-Anchored Adaptive Local Token Set}
\label{subsec:adaptive_local_set}

Let \(d_{\min}\) and \(d_{\max}\) denote the minimum and maximum local-set
sizes, with \(2\leq d_{\min}\leq d_{\max}<|\V|\). For each
\(u\in\Tbatch\), the student first proposes the \(d_{\max}\) vocabulary tokens
with the largest probabilities:
\begin{equation}
    \Sset_u^{\max}
    =
    \TopD(P_u^S,d_{\max}),
    \qquad
    a_u
    =
    \arg\max_{i\in\V}p_{u,i}^T,
    \label{eq:proposal_and_teacher_top1}
\end{equation}
where \(\TopD(P,d)\) returns the vocabulary indices of the \(d\) largest
components of \(P\), and \(a_u\) is the teacher top-1 token. The set
\(\Sset_u^{\max}\) is the \emph{student proposal set}; it is not the final
local token set. Because the student's ranking may be unreliable, especially
early in training, ALRA forms a teacher-anchored candidate set of the same
cardinality:
\begin{equation}
    \C_u^{\max}
    =
    \begin{cases}
    \Sset_u^{\max},
    & a_u\in\Sset_u^{\max},\\[2pt]
    \TopD(P_u^S,d_{\max}-1)\cup\{a_u\},
    & a_u\notin\Sset_u^{\max}.
    \end{cases}
    \label{eq:teacher_anchored_candidate_set}
\end{equation}
Thus, \(|\C_u^{\max}|=d_{\max}\). When \(a_u\notin\Sset_u^{\max}\), the
lowest-ranked token in the original \(d_{\max}\)-element student proposal is
replaced by the teacher top-1 token. This operation preserves a candidate set
largely based on the student proposal while ensuring that it contains at least
one teacher-preferred reference. The teacher top-1 token is not treated as a
ground-truth label.

The candidate set is intended to capture the tokens that the student currently
considers most likely. We therefore construct most of the set from the tokens
assigned the highest probabilities by the student. A teacher-only candidate
set may exclude tokens that receive high student probability and therefore
may not reflect the student's current prediction state.

The teacher top-1 token is included as a minimal anchor. It ensures that the
teacher's most probable token remains available for local supervision when it
is absent from the student proposal, while replacing at most one
student-proposed token and keeping the candidate-set size fixed. Including
multiple teacher-ranked tokens would introduce more teacher-selected tokens
into the fixed candidate budget and make the candidate set less responsive to
the student's current predictions. It would also require an additional rule
for combining the teacher and student proposals within the fixed budget. ALRA
therefore uses only the teacher top-1 token as an anchor, while the teacher
probabilities within the anchored candidate set are used in the following
steps to compute the effective support size, determine the position-specific
local budget, and rank the final local tokens.

ALRA next measures the dispersion of the teacher distribution conditioned on
the anchored candidate set \(\C_u^{\max}\). Specifically, it restricts the
teacher distribution to \(\C_u^{\max}\), renormalizes the retained
probabilities, and computes the conditional entropy and corresponding
effective support size:
\begin{equation}
\begin{aligned}
    \rho_{u,i}
    &=
    \frac{p_{u,i}^T}
    {\sum_{j\in\C_u^{\max}}p_{u,j}^T},
    && i\in\C_u^{\max},\\
    H_u^{\mathrm{loc}}
    &=
    -\sum_{i\in\C_u^{\max}}
    \rho_{u,i}\log\rho_{u,i},
    \qquad
    E_u^{\mathrm{loc}}
    =
    \exp(H_u^{\mathrm{loc}}).
\end{aligned}
\label{eq:candidate_conditioned_teacher_support}
\end{equation}
Here, \(H_u^{\mathrm{loc}}\) is the teacher conditional entropy within the
teacher-anchored candidate set; it is not the teacher's full-vocabulary
entropy. The quantity \(E_u^{\mathrm{loc}}\), obtained by exponentiating this
entropy, is the corresponding effective support size. A concentrated
conditional distribution yields a smaller effective support size, whereas a
more diffuse distribution yields a larger one, indicating that the teacher
probability is distributed across a larger effective number of candidate
alternatives.

The batch-average effective support size over the valid prediction positions
in the current forward batch is
\begin{equation}
    \bar{E}_{\Tbatch}^{\mathrm{loc}}
    =
    \frac{1}{|\Tbatch|}
    \sum_{v\in\Tbatch}
    E_v^{\mathrm{loc}}.
    \label{eq:batch_average_effective_support}
\end{equation}
ALRA then assigns each prediction position an integer local-set size:
\begin{equation}
    d_u
    =
    \clip\!\left(
    \round\!\left[
    d_{\min}
    +
    (d_{\max}-d_{\min})
    \frac{E_u^{\mathrm{loc}}}
    {\bar{E}_{\Tbatch}^{\mathrm{loc}}+\epsilon}
    \right],
    d_{\min},d_{\max}
    \right),
    \label{eq:adaptive_local_budget}
\end{equation}
where \(\epsilon>0\) avoids division by zero. The ratio
\(E_u^{\mathrm{loc}}/(\bar{E}_{\Tbatch}^{\mathrm{loc}}+\epsilon)\) compares
the candidate-conditioned teacher effective support size at position \(u\)
with its average over the valid prediction positions in the current forward
batch. Rounding converts the resulting continuous budget to an integer, while
clipping enforces the prescribed bounds
\(d_{\min}\leq d_u\leq d_{\max}\).

This batch-relative normalization allows the local budget to vary across
prediction positions and to adapt as the student proposals evolve during
training. Early in training, the student proposal may contain tokens that
receive little teacher probability. After insertion of the teacher top-1
token, the resulting candidate-conditioned teacher distribution may therefore
be strongly concentrated around the anchor, yielding small values of
\(E_u^{\mathrm{loc}}\). When this behavior is common across the current batch,
\(\bar{E}_{\Tbatch}^{\mathrm{loc}}\) is also small. Positions whose effective
support is close to the batch average therefore have a ratio near one and can
receive large, often maximum, local budgets. This provides broad local
supervision while the student ranking may still be unreliable.

As training progresses, the student proposal may include more tokens that
receive substantial probability under the teacher distribution. At positions
where the teacher distributes probability mass across multiple candidate
tokens, the candidate-conditioned teacher distribution becomes more diffuse,
yielding a larger effective support size. The batch-relative ratio is
therefore larger at these positions than at positions with more concentrated
candidate-conditioned teacher distributions. After rounding and clipping,
positions with larger effective support relative to the batch average tend to
receive larger local sets, including more competing tokens for supervision,
whereas positions with smaller relative effective support receive smaller
local sets. Thus, \(d_u\) depends on both the candidate-conditioned teacher
distribution at position \(u\) and the batch-average effective support over
the valid positions in the same forward pass.
Fig.~\ref{fig:budget_dynamics} empirically illustrates the budget assignments
produced by Eq.~\eqref{eq:adaptive_local_budget} at different stages of
training.

Finally, ALRA selects the \(d_u\) highest-probability tokens under the teacher
distribution within the anchored candidate set:
\begin{equation}
    \I_u
    =
    \TopD\!\left(
    \{p_{u,i}^T:i\in\C_u^{\max}\},d_u
    \right),
    \qquad
    \R_u
    =
    \V\setminus\I_u.
    \label{eq:adaptive_local_partition}
\end{equation}
When \(\TopD\) is applied to values indexed by \(\C_u^{\max}\), it returns
the corresponding vocabulary indices from that set. Therefore, \(\I_u\) is
the final position-specific adaptive local token set, while \(\R_u\) is its
full-vocabulary complement. Tokens that are not selected into the local set
are not discarded. They remain in the rest region and are still supervised
through the rest-conditional term of ALD introduced in the next subsection.

\subsection{Adaptive Local Divergence}
\label{subsec:ald}

Given the adaptive partition \((\I_u,\R_u)\), we first define the teacher and
student probability masses assigned to the local and rest regions, together
with the corresponding binary region distributions:
\begin{equation}
\begin{aligned}
    \alpha_u^T
    &=
    \sum_{i\in\I_u}p_{u,i}^T,
    &
    \bar{\alpha}_u^T
    &=
    1-\alpha_u^T,
    &
    b_u^T
    &=
    (\alpha_u^T,\bar{\alpha}_u^T),\\
    \alpha_u^S
    &=
    \sum_{i\in\I_u}p_{u,i}^S,
    &
    \bar{\alpha}_u^S
    &=
    1-\alpha_u^S,
    &
    b_u^S
    &=
    (\alpha_u^S,\bar{\alpha}_u^S).
\end{aligned}
\label{eq:local_rest_masses}
\end{equation}
Here, \(\alpha_u^T\) and \(\alpha_u^S\) are the total probabilities assigned
to the local region \(\I_u\), whereas \(\bar{\alpha}_u^T\) and
\(\bar{\alpha}_u^S\) are the corresponding probabilities assigned to its
complement \(\R_u\).

Within each region, the retained token probabilities are renormalized to
define conditional distributions. For each token \(i\in\I_u\), the teacher
and student conditional probabilities are
\begin{equation}
\begin{aligned}
    \tilde{p}_{u,i}^{T,\I}
    &=
    \frac{p_{u,i}^T}{\alpha_u^T},
    &
    \tilde{p}_{u,i}^{S,\I}
    &=
    \frac{p_{u,i}^S}{\alpha_u^S}.
\end{aligned}
\label{eq:local_conditional_probabilities}
\end{equation}
These probabilities define the teacher and student conditional distributions
\(\tilde{P}_u^{T,\I}\) and \(\tilde{P}_u^{S,\I}\) over the local region
\(\I_u\).

For each token \(i\in\R_u\), the corresponding conditional probabilities in
the rest region are
\begin{equation}
\begin{aligned}
    \tilde{p}_{u,i}^{T,\R}
    &=
    \frac{p_{u,i}^T}{\bar{\alpha}_u^T},
    &
    \tilde{p}_{u,i}^{S,\R}
    &=
    \frac{p_{u,i}^S}{\bar{\alpha}_u^S}.
\end{aligned}
\label{eq:rest_conditional_probabilities}
\end{equation}
These probabilities define the teacher and student conditional distributions
\(\tilde{P}_u^{T,\R}\) and \(\tilde{P}_u^{S,\R}\) over the rest region
\(\R_u\). By construction,
\[
    \sum_{i\in\I_u}\tilde{p}_{u,i}^{T,\I}
    =
    \sum_{i\in\I_u}\tilde{p}_{u,i}^{S,\I}
    =
    1,
    \qquad
    \sum_{i\in\R_u}\tilde{p}_{u,i}^{T,\R}
    =
    \sum_{i\in\R_u}\tilde{p}_{u,i}^{S,\R}
    =
    1.
\]
Thus, the local distributions are conditioned on the next token belonging to
\(\I_u\), while the rest distributions are conditioned on the next token
belonging to \(\R_u\).

The full-vocabulary forward KL can be decomposed exactly as
\begin{equation}
    \KL(P_u^T\|P_u^S)
    =
    \KL(b_u^T\|b_u^S)
    +
    \alpha_u^T\KL(\tilde{P}_u^{T,\I}\|\tilde{P}_u^{S,\I})
    +
    \bar{\alpha}_u^T\KL(\tilde{P}_u^{T,\R}\|\tilde{P}_u^{S,\R}).
    \label{eq:exact_local_rest_decomposition}
\end{equation}

Equation~\eqref{eq:exact_local_rest_decomposition} shows that the
full-vocabulary forward KL consists of three parts: a binary mass-matching
term, a local-conditional divergence weighted by \(\alpha_u^T\), and a
rest-conditional divergence weighted by \(\bar{\alpha}_u^T\).

The coefficients \(\alpha_u^T\) and \(\bar{\alpha}_u^T\) are the teacher
probability masses assigned to the local and rest regions, respectively. They
arise directly from the exact KL decomposition and are not tunable
hyperparameters. Retaining these coefficients preserves equality with the
original full-vocabulary forward KL, but also scales each conditional
divergence in proportion to the teacher probability mass of its region.
Consequently, when one region has small teacher mass, the contribution of its
conditional divergence to the exact KL objective is reduced by the
corresponding mass coefficient. Thus, for a fixed conditional discrepancy,
that region contributes less than it would under a unit-weighted conditional
objective.

This effect is particularly relevant to the rest region when the adaptive
local set captures most of the teacher probability mass. In that case,
\(\bar{\alpha}_u^T\) is small, so the contribution of the rest-conditional
divergence is reduced in the exact decomposition. Although this coefficient is
mathematically required to recover the original full-vocabulary KL, it gives
the rest-conditional divergence less weight than an objective that assigns
this term a unit coefficient.

We therefore define Adaptive Local Divergence (ALD) by retaining the
mass-matching term while assigning unit coefficients to the two conditional
divergences:
\begin{equation}
\begin{aligned}
    \mathcal{L}_{\mathrm{ALD}}(u)
    &=
    \mathcal{L}_{\mathrm{mass}}(u)
    +
    \mathcal{L}_{\mathrm{local}}(u)
    +
    \mathcal{L}_{\mathrm{rest}}(u),\\
    \mathcal{L}_{\mathrm{mass}}(u)
    &=
    \KL(b_u^T\|b_u^S),\\
    \mathcal{L}_{\mathrm{local}}(u)
    &=
    \KL(\tilde{P}_u^{T,\I}\|\tilde{P}_u^{S,\I}),\\
    \mathcal{L}_{\mathrm{rest}}(u)
    &=
    \KL(\tilde{P}_u^{T,\R}\|\tilde{P}_u^{S,\R}).
\end{aligned}
\label{eq:ald}
\end{equation}

The mass term aligns the total teacher and student probability assigned to the
local and rest regions. The local-conditional term aligns their relative
probability distributions within the adaptive local set, while the
rest-conditional term aligns the corresponding relative distributions over
the complementary vocabulary region. Assigning unit coefficients to the two
conditional terms prevents either term from being downweighted solely because
its region has small teacher probability mass.

This design does not imply that the three terms have similar scales or produce
equal gradient contributions. Rather, it removes the explicit teacher-mass
scaling from the two conditional divergences and allows both conditional terms
to contribute without teacher-mass scaling. The unit coefficients also avoid
introducing additional region-level weighting hyperparameters.

Because ALD removes the probability-mass coefficients from the conditional
terms, \(\mathcal{L}_{\mathrm{ALD}}(u)\) is generally not algebraically
identical to \(\KL(P_u^T\|P_u^S)\). ALD is therefore a distinct region-aware
objective that uses the three components of the exact local--rest KL
decomposition, rather than an exact rewriting of the original forward KL. The
exact decomposition is derived in Appendix~\ref{app:kl_decomposition}.

\subsection{Student-Weighted Pairwise Relational Alignment}
\label{subsec:pairwise_alignment}

The local-conditional term in ALD aligns the complete teacher and student
conditional distributions on \(\I_u\), but it does not assign separate
importance to discrepancies between individual local token pairs. Pairwise
relational distillation provides an explicit two-token view of relative
preference \citep{xu2025local}. In contrast to image classification,
where the class set is fixed across samples, ALRA defines its pairwise
relations over the position-specific local token set \(\I_u\). Moreover,
rather than using class-rank-based weights, ALRA derives each pair weight from
the student's full-vocabulary probabilities.

Given \(|\I_u|=d_u\), ALRA constructs the set of all unordered pairs of local
vocabulary indices:
\begin{equation}
    \Pset_u
    =
    \{(i,j):i,j\in\I_u,\ i<j\},
    \qquad
    |\Pset_u|
    =
    \binom{d_u}{2}.
    \label{eq:pair_set}
\end{equation}
For each \((i,j)\in\Pset_u\), the corresponding teacher and student two-token
distributions are
\begin{equation}
\begin{aligned}
    r_u^T(i,j)
    &=
    \softmax\!\left(
    [z_{u,i}^T/\tau_p,\ z_{u,j}^T/\tau_p]
    \right),\\
    r_u^S(i,j)
    &=
    \softmax\!\left(
    [z_{u,i}^S/\tau_p,\ z_{u,j}^S/\tau_p]
    \right),
\end{aligned}
\label{eq:pairwise_two_token_distribution}
\end{equation}
where \(\tau_p>0\) is the pairwise temperature. These two-way softmax
distributions represent the teacher's and student's relative preference
between the two vocabulary tokens.

A general weighted pairwise relational objective is
\begin{equation}
    \mathcal{L}_{\mathrm{pair}}^{\phi}(u)
    =
    \sum_{(i,j)\in\Pset_u}
    \kappa_{u,ij}^{\phi}
    \KL\!\left(
    r_u^T(i,j)\|r_u^S(i,j)
    \right),
    \label{eq:general_pairwise_loss}
\end{equation}
where \(\phi\) denotes the weighting scheme and
\(\kappa_{u,ij}^{\phi}\) is the weight assigned to pair \((i,j)\). The KL
direction is teacher-to-student for every pair.

For comparison, direct Pairwise Relational Alignment (PRA) uses unit pair
weights, while PRA-U uses normalized uniform weights:
\begin{equation}
\begin{aligned}
    \kappa_{u,ij}^{\mathrm{R}}
    &=1,
    &
    \mathcal{L}_{\mathrm{PRA}}(u)
    &=
    \mathcal{L}_{\mathrm{pair}}^{\mathrm{R}}(u),\\
    \kappa_{u,ij}^{\mathrm{U}}
    &=
    \frac{1}{|\Pset_u|},
    &
    \mathcal{L}_{\mathrm{PRA\text{-}U}}(u)
    &=
    \mathcal{L}_{\mathrm{pair}}^{\mathrm{U}}(u),
    \qquad (i,j)\in\Pset_u.
\end{aligned}
\label{eq:pra_variants}
\end{equation}
PRA is unnormalized, so its loss scale can vary with the number of local
pairs, whereas PRA-U has unit total pair weight. This difference is relevant
when comparing variants with different local-set sizes.

For SWPRA, ALRA first computes an unnormalized score for each local token
pair:
\begin{equation}
    s_{u,ij}
    =
    \exp\!\left(
    -\gamma |p_{u,i}^S-p_{u,j}^S|
    \right)
    (p_{u,i}^S+p_{u,j}^S),
    \qquad
    (i,j)\in\Pset_u,
    \label{eq:unnormalized_pair_score}
\end{equation}


\begin{figure}[pos=!t]
    \centering

    \begin{minipage}[t]{0.305\textwidth}
        \centering
        \includegraphics[
            width=\linewidth,
            keepaspectratio
        ]{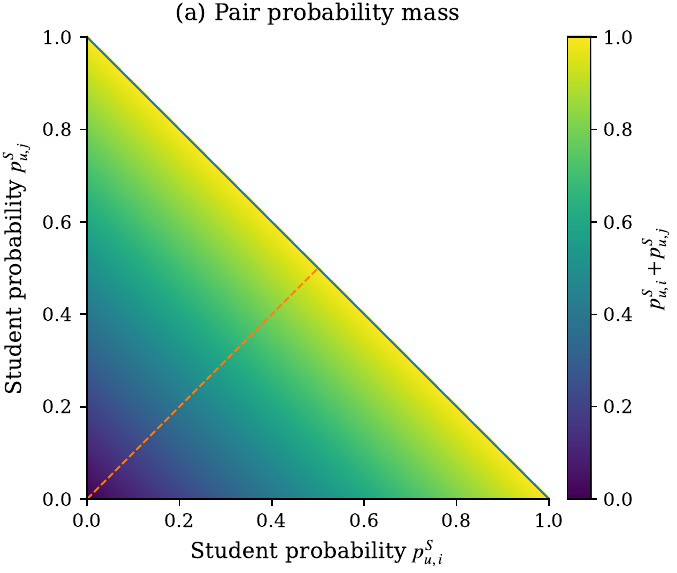}
    \end{minipage}
    \hfill
    \begin{minipage}[t]{0.305\textwidth}
        \centering
        \includegraphics[
            width=\linewidth,
            keepaspectratio
        ]{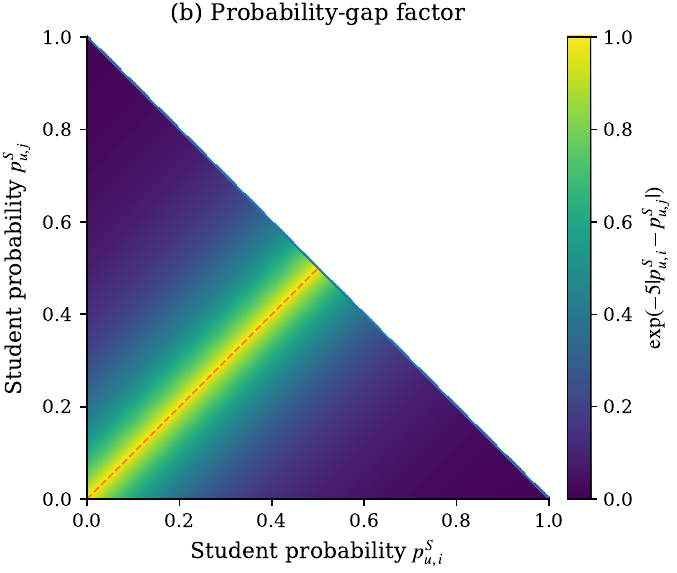}
    \end{minipage}
    \hfill
    \begin{minipage}[t]{0.305\textwidth}
        \centering
        \includegraphics[
            width=\linewidth,
            keepaspectratio
        ]{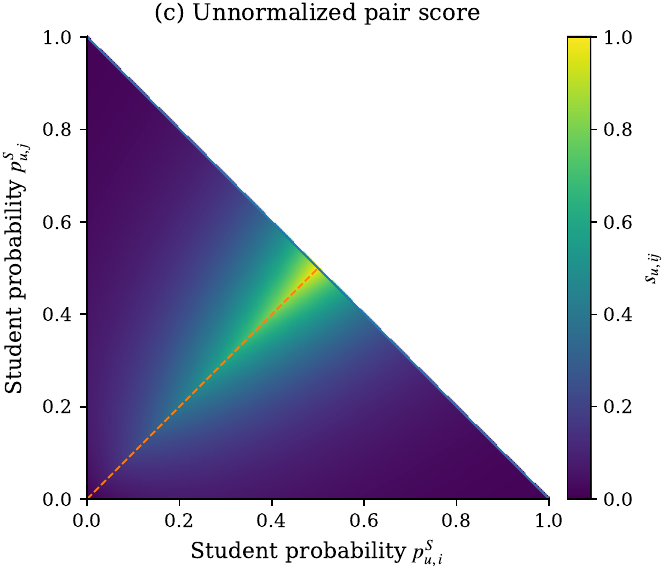}
    \end{minipage}

    \caption{
    Pair-score construction for Student-Weighted Pairwise Relational Alignment
    with \(\gamma=5\), as used in the main experiments.
    (a) Pair probability mass \(p_{u,i}^S+p_{u,j}^S\).
    (b) Probability-gap factor
    \(\exp[-5|p_{u,i}^S-p_{u,j}^S|]\).
    (c) Unnormalized pair score \(s_{u,ij}\), obtained by multiplying the two
    factors. The scores are subsequently rescaled over \(\Pset_u\) to obtain
    the SWPRA pair weights \(\omega_{u,ij}\). The region
    \(p_{u,i}^S+p_{u,j}^S>1\) is excluded because the two probabilities belong
    to the same full-vocabulary distribution. The dashed diagonal marks
    \(p_{u,i}^S=p_{u,j}^S\), where the probability gap is zero.
    }
    \label{fig:swpra_weighting}
\end{figure}


where \(\gamma>0\) controls the sensitivity to the student probability gap.
The corresponding SWPRA pair weight is
\begin{equation}
    \kappa_{u,ij}^{\mathrm{SW}}
    =
    \omega_{u,ij}
    =
    \frac{s_{u,ij}}
    {\sum_{(a,b)\in\Pset_u}s_{u,ab}+\epsilon},
    \qquad
    (i,j)\in\Pset_u.
    \label{eq:normalized_pair_weight}
\end{equation}

The factor \(p_{u,i}^S+p_{u,j}^S\) measures the total probability mass of the
pair under the student's original full-vocabulary distribution, rather than
under the two-token renormalization. The factor
\(\exp[-\gamma|p_{u,i}^S-p_{u,j}^S|]\) decreases as the student probability
gap increases. Their product therefore favors pairs that carry substantial
student probability mass and have similar probabilities under the student
distribution. A pair containing two low-probability tokens receives a small
score even when its probability gap is small, while a high-mass pair is
downweighted when its two probabilities are widely separated.

The denominator in Eq.~\eqref{eq:normalized_pair_weight} rescales the scores
by their sum over \(\Pset_u\), while \(\epsilon>0\) prevents numerical issues
when that sum is very small. The resulting weight \(\omega_{u,ij}\)
determines the relative contribution of pair \((i,j)\) to the pairwise
objective. These weights multiply the corresponding teacher-to-student
pairwise KL terms and do not change the teacher pairwise targets. The
probabilities used to compute the scores and weights are taken from \(P_u^S\)
at temperature \(\tau\), whereas the relational distributions in
Eq.~\eqref{eq:pairwise_two_token_distribution} use \(\tau_p\). Fig.~\ref{fig:swpra_weighting} separates the two factors in
Eq.~\eqref{eq:unnormalized_pair_score} and shows how their product forms the
unnormalized pair score. The figure uses \(\gamma=5\), matching the value used
in the main experiments.






The Student-Weighted Pairwise Relational Alignment loss is
\begin{equation}
    \mathcal{L}_{\mathrm{SWPRA}}(u)
    =
    \mathcal{L}_{\mathrm{pair}}^{\mathrm{SW}}(u).
    \label{eq:swpra_loss}
\end{equation}
PRA, PRA-U, and SWPRA use the same teacher-to-student KL divergence for each
two-token pair and differ only in how the pairwise terms are weighted.

\subsection{Training Objective}
\label{subsec:objective}

The ALRA distillation objective over the current forward batch is
\begin{equation}
    \mathcal{L}_{\mathrm{ALRA}}
    =
    \frac{1}{|\Tbatch|}
    \sum_{u\in\Tbatch}
    \left[
    \mathcal{L}_{\mathrm{ALD}}(u)
    +
    \lambda_{\mathrm{pair}}
    \mathcal{L}_{\mathrm{SWPRA}}(u)
    \right],
    \label{eq:overall_alra_loss}
\end{equation}
where \(\lambda_{\mathrm{pair}}\geq0\) controls the contribution of the
pairwise relational term. The mass term is retained unchanged, while the
local-conditional and rest-conditional components of
\(\mathcal{L}_{\mathrm{ALD}}\) use unit coefficients, as defined in
Eq.~\eqref{eq:ald}; no additional region-level weighting hyperparameters are
introduced. In the comparison variants, \(\mathcal{L}_{\mathrm{SWPRA}}\) is
replaced by \(\mathcal{L}_{\mathrm{PRA}}\) or
\(\mathcal{L}_{\mathrm{PRA\text{-}U}}\).

When ground-truth next-token labels are available, we define the student's
temperature-one next-token distribution \(Q_u^S=\softmax(z_u^S)\), with
component \(q_{u,i}^S\), and optimize
\begin{equation}
    \mathcal{L}_{\mathrm{train}}
    =
    \mathcal{L}_{\mathrm{ALRA}}
    +
    \lambda_{\mathrm{CE}}
    \frac{1}{|\Tbatch|}
    \sum_{u\in\Tbatch}
    -\log q_{u,y_u}^S,
    \label{eq:training_loss_with_ce}
\end{equation}
where \(y_u\) is the ground-truth next-token label associated with prediction
position \(u\), and \(\lambda_{\mathrm{CE}}\geq0\) controls the causal
language-modeling term.

Algorithm~\ref{alg:alra} shows how ALRA is computed within one forward batch.
Line~1 obtains the teacher and student distributions for all valid prediction
positions. Lines~2--6 build the teacher-anchored candidate set at each position
and compute its candidate-conditioned teacher effective support. Line~7
averages these values over the current batch. Lines~8--15 then use this batch
statistic to determine the local budget, form the local--rest partition, and
compute the ALD and SWPRA terms at each position. Finally, Line~16 averages
the position-level objectives to obtain
\(\mathcal{L}_{\mathrm{ALRA}}\). When the causal language-modeling term is
used, it is added afterward as defined in
Eq.~\eqref{eq:training_loss_with_ce}.

\begin{algorithm}[!t]
\caption{Adaptive Local Relational Alignment}
\label{alg:alra}
\begin{algorithmic}[1]
\Require Valid next-token prediction positions \(\Tbatch\) from the current
forward batch; teacher logits \(\{z_u^T\}_{u\in\Tbatch}\); student logits
\(\{z_u^S\}_{u\in\Tbatch}\); local-set bounds \(d_{\min},d_{\max}\);
temperatures \(\tau,\tau_p\); probability-gap sensitivity \(\gamma\);
pairwise-loss coefficient \(\lambda_{\mathrm{pair}}\).
\State Compute \(P_u^T=\softmax(z_u^T/\tau)\) and
\(P_u^S=\softmax(z_u^S/\tau)\) for all \(u\in\Tbatch\).
\For{\(u\in\Tbatch\)}
    \State Compute \(\Sset_u^{\max}\) and \(a_u\) using
    Eq.~\eqref{eq:proposal_and_teacher_top1}.
    \State Construct \(\C_u^{\max}\) using
    Eq.~\eqref{eq:teacher_anchored_candidate_set}.
    \State Compute \(E_u^{\mathrm{loc}}\) using
    Eq.~\eqref{eq:candidate_conditioned_teacher_support}.
\EndFor
\State Compute \(\bar{E}_{\Tbatch}^{\mathrm{loc}}\) using
Eq.~\eqref{eq:batch_average_effective_support}.
\For{\(u\in\Tbatch\)}
    \State Compute \(d_u\) using Eq.~\eqref{eq:adaptive_local_budget}.
    \State Construct \((\I_u,\R_u)\) using
    Eq.~\eqref{eq:adaptive_local_partition}.
    \State Compute \(\mathcal{L}_{\mathrm{ALD}}(u)\) using
    Eq.~\eqref{eq:ald}.
    \State Construct \(\Pset_u\) using Eq.~\eqref{eq:pair_set}.
    \State Compute \(s_{u,ij}\) and \(\omega_{u,ij}\) using
    Eqs.~\eqref{eq:unnormalized_pair_score}--%
    \eqref{eq:normalized_pair_weight}.
    \State Compute \(\mathcal{L}_{\mathrm{SWPRA}}(u)\) using
    Eq.~\eqref{eq:swpra_loss}.
\EndFor
\State \Return \(\mathcal{L}_{\mathrm{ALRA}}\) using
Eq.~\eqref{eq:overall_alra_loss}.
\end{algorithmic}
\end{algorithm}

\FloatBarrier

\section{Experiments}
\label{sec:experiments}

\subsection{Experimental Setup}
\label{subsec:experimental_setup}
We conduct pre-training distillation with a fixed Qwen1.5-1.8B teacher and Qwen-architecture students using a processed corpus constructed from Pile Uncopyrighted, a filtered version of The Pile. All students are initialized randomly and trained from scratch. Within each student setting, all compared methods use the same processed data in the same order, sequence length, assigned token budget, and shared optimization schedule. For each method, we report the checkpoint at the end of the assigned token budget; the full-budget runs are evaluated zero-shot on nine held-out downstream benchmarks.

\paragraph{Models and initialization.}
We use Qwen1.5-1.8B\footnote{\url{https://huggingface.co/Qwen/Qwen1.5-1.8B}} as the fixed teacher and instantiate two Qwen student configurations corresponding to Pretrain-Qwen-200M\footnote{\url{https://huggingface.co/MiniLLM/Pretrain-Qwen-200M}} and Pretrain-Qwen-500M\footnote{\url{https://huggingface.co/MiniLLM/Pretrain-Qwen-500M}}. We adopt these architectural configurations from prior Qwen-based pre-training experiments \citep{gu2024miniplm}, but do not initialize from the released student weights: both students are initialized randomly and optimized from scratch, while the pretrained teacher remains frozen. For each student architecture, all compared methods use the same fixed random seed for model initialization and therefore start from the same initial parameter values. This controls for variation due to random initialization when comparing distillation objectives. Teacher and students share the same tokenizer and vocabulary, so their output probabilities are aligned token-wise and no cross-tokenizer vocabulary mapping is required. The two students provide two capacity regimes under the same training protocol, allowing us to assess whether the observed trends persist across student scale. Their exact architectural configurations are reported in Appendix~\ref{app:model_configurations} and Table~\ref{tab:model_configurations}.

\paragraph{Pre-training data and budgets.}
The pre-training corpus is constructed from Pile
Uncopyrighted\footnote{\url{https://huggingface.co/datasets/monology/pile-uncopyrighted}},
a filtered version of The Pile \citep{gao2020pile,biderman2022datasheet}
obtained by removing the Books3, BookCorpus2, OpenSubtitles, YTSubtitles,
and OWT2 subsets. We publicly release the exact processed training data used in our experiments. All compared methods within one student setting observe the same processed examples in the same order. The main baseline comparison and the pairwise weighting study use the full-budget run of approximately 1.035B nominal model-input tokens (126.4K optimizer updates), which we refer to as the approximately 1B-token setting. The fixed-budget and candidate-set anchoring studies instead terminate after approximately 550.5M nominal model-input tokens (67.2K optimizer updates), using the corresponding prefix of the same processed training stream. These shorter runs are used only for controlled mechanism analyses; their checkpoints are never mixed into the nine-benchmark held-out comparison, so every held-out result is reported at the same approximately 1B-token budget. Source-document selection, tokenization, boundary-aware fragment construction, exact token-budget accounting, and the relation between the stored 513-token examples and the 512-token model input are described in Appendix~\ref{app:training_data}.

\paragraph{Compared methods.}
We compare ALRA against eight baselines: Pre-train w/o KD, Vanilla KD~\citep{muralidharan2024compact}, PD~\citep{peng2025pre}, ATKD~\citep{zhong2024revisiting}, RLD~\citep{sun2025knowledge}, LDRLD~\citep{xu2025local}, TAD~\citep{dasguptadon}, and BiLD~\citep{li2025bild}. Pre-train w/o KD denotes standard causal language model pre-training without distillation. In our implementation, Vanilla KD combines the causal language modeling objective with a full-vocabulary forward-KL loss between teacher and student distributions. PD denotes our adaptation of a pre-training distillation configuration proposed by \citep{peng2025pre}. ATKD uses teacher uncertainty to estimate token difficulty and applies different teaching modes to easy- and hard-to-learn tokens. RLD uses ground-truth label information to dynamically refine teacher logits while preserving class correlations, whereas LDRLD (Local Dense Relational Logit Distillation) captures fine-grained inter-class logit relations through recursive decoupling and recombination with adaptive pairwise weighting. TAD (Tail-Aware Distillation) decouples the teacher's top-\(K\) probabilities from the remaining tail to reduce mode dominance and increase the contribution of lower-probability predictions, whereas BiLD (Bi-directional Logits Difference) uses teacher-led and student-led logit-difference matching to exploit internal logit-ranking information while filtering long-tail noise.

Several of these methods were originally introduced under different training regimes or output spaces. We therefore distinguish each method's defining objective from our from-scratch autoregressive adaptation and document the exact implementation used in Appendix~\ref{app:baseline_implementations}. Within each student setting, the teacher--student pair, processed data and data order, token budget, sequence length, and optimizer schedule are held fixed; method-specific loss constructions and associated hyperparameters vary as required. The comparisons are therefore matched by student update and token budget rather than by total training FLOPs; runtime and memory overheads are reported separately in Appendix~\ref{app:hardware}.

\paragraph{Evaluation protocol.}
We perform zero-shot evaluation on nine downstream tasks widely used to assess the zero-shot capabilities of base language models~\citep{llama2023llama,groeneveld2024olmo,gu2024miniplm}, using the latest commit of the LM Evaluation Harness~\citep{gao2021framework} available at the time of evaluation. The held-out suite contains HellaSwag (HS)~\citep{zellers2019hellaswag}, LAMBADA-OpenAI (LAMB)~\citep{paperno2016lambada}, WinoGrande (WG)~\citep{sakaguchi2021winogrande}, OpenBookQA (OBQA)~\citep{mihaylov2018can}, ARC-Challenge and ARC-Easy (ARC-C/ARC-E)~\citep{clark2018think}, PIQA~\citep{bisk2020piqa}, SocialIQA (SIQA)~\citep{sap2019social}, and StoryCloze-2016 (SC16)~\citep{mostafazadeh2016corpus}. No labeled downstream examples are used to update the student parameters.

To guarantee a labeled held-out set for every task, we assign it by a fixed rule: we use the official test split where its public labels are usable (LAMBADA-OpenAI, OpenBookQA, ARC-Easy, ARC-Challenge, StoryCloze-2016) and otherwise fall back to the public validation split (HellaSwag, WinoGrande, PIQA, SocialIQA). A separate validation set, used only for the controlled 550M-token analyses and never for gradient-based task adaptation, is available for HellaSwag, ARC-Easy, and PIQA. No separate validation set is used for the remaining tasks. This protocol keeps all models under comparable zero-shot conditions, with no fine-tuning on labeled data. The exact task identifiers, the evaluation-file sizes, and the full split-assignment strategy are reported in Appendix~\ref{app:evaluation_details}, Table~\ref{tab:evaluation_suite}, and Fig.~\ref{fig:evaluation_sizes}. ``Avg.'' is an unweighted arithmetic mean, so every benchmark contributes equally regardless of its number of examples.

\paragraph{Training and checkpoint reporting.}
All students traverse their assigned budget-limited prefix of the processed training stream once, using the shared optimization configuration in Appendix~\ref{app:optimization} and Table~\ref{tab:optimization_configuration}. For each method, we report the final checkpoint at the end of its assigned token budget rather than selecting the best checkpoint based on validation or held-out performance; downstream scores therefore do not influence checkpoint selection. ALRA uses \(d_{\min}=3\) and \(d_{\max}=25\); the remaining ALRA hyperparameters are listed in Table~\ref{tab:alra_hyperparameters}. Because the reported comparisons do not include across-seed statistics, we do not claim statistical significance for small numerical differences.

\subsection{Main Results on Held-Out Benchmarks}
\label{subsec:main_results}

\begin{table}[!tbp]
\caption{
Held-out zero-shot accuracy (\%) of randomly initialized 200M- and
500M-parameter students trained under an approximately 1B-token budget.
Qwen1.5-1.8B is used as the teacher for all KD methods, while
Pre-Train w/o KD uses only the causal language-modeling objective.
The compared KD methods are Vanilla KD
\basecite{muralidharan2024compact}, PD
\basecite{peng2025pre}, ATKD
\basecite{zhong2024revisiting}, RLD
\basecite{sun2025knowledge}, LDRLD
\basecite{xu2025local}, TAD
\basecite{dasguptadon}, and BiLD
\basecite{li2025bild}. ``Avg.'' is the arithmetic mean over ARC-Challenge,
ARC-Easy, HellaSwag, LAMBADA-OpenAI, OpenBookQA, PIQA, SocialIQA,
StoryCloze-2016, and Winogrande.
$\Delta_{\mathrm{Avg.}}\!\uparrow$ is the absolute difference from
Pre-Train w/o KD under the same student setting, computed from unrounded
averages. Within each student setting, the best and second-best benchmark
values are shown in bold and underlined, respectively. The best average
and largest improvement are shown in bold dark red. Ties receive the same
marking.
}
\label{tab:baseline_comparison}
\centering
\scriptsize
\setlength{\tabcolsep}{2.8pt}
\renewcommand{\arraystretch}{1.12}
\begin{adjustbox}{max width=\textwidth}
\begin{tabular}{@{}llcccccccccccc@{}}
\toprule
\multirow{2}{*}{\textbf{Teacher/Student}} &
\multirow{2}{*}{\textbf{Method}} &
\multicolumn{1}{c}{\textbf{Publication}} &
\multirow{2}{*}{\textbf{ARC-C}} &
\multirow{2}{*}{\textbf{ARC-E}} &
\multirow{2}{*}{\textbf{HS}} &
\multirow{2}{*}{\textbf{LAMB}} &
\multirow{2}{*}{\textbf{OBQA}} &
\multirow{2}{*}{\textbf{PIQA}} &
\multirow{2}{*}{\textbf{SIQA}} &
\multirow{2}{*}{\textbf{SC16}} &
\multirow{2}{*}{\textbf{WG}} &
\multirow{2}{*}{\textbf{Avg.}} &
\multirow{2}{*}{$\boldsymbol{\Delta}_{\mathrm{Avg.}}\!\uparrow$} \\
\cmidrule(lr){3-3}
& & \textbf{Year} & & & & & & & & & & & \\
\midrule
\multirow{9}{*}{\makecell[c]{Qwen1.5\\1.8B\\$\downarrow$\\200M}}
& Pre-Train w/o KD & \pubyear{--}
& 21.84 & 32.45 & 26.46 & 12.92 & 23.80 & 55.28 & 34.03 & 51.50 & 50.51 & 34.31 & Ref. \\
& Vanilla KD & \pubyear{NeurIPS'24}
& 21.50 & 32.91 & 27.25 & 16.69 & 24.40 & 56.37 & \tabsecond{35.21} & 52.50 & 49.33 & 35.13 & +0.82 \\
& PD & \pubyear{ACL'25}
& 19.88 & 33.80 & 26.09 & \tabbest{21.06} & 26.00 & 56.64 & 34.85 & 52.10 & 49.96 & 35.60 & +1.29 \\
& ATKD & \pubyear{ACL'24}
& \tabsecond{23.46} & 34.81 & 27.36 & 15.97 & 24.20 & 58.11 & 34.85 & 52.00 & 48.78 & 35.50 & +1.19 \\
& RLD & \pubyear{ICCV'25}
& 22.10 & 35.06 & \tabsecond{27.73} & 12.75 & 26.00 & 57.34 & 34.19 & 52.30 & \tabsecond{51.78} & 35.47 & +1.16 \\
& LDRLD & \pubyear{ICCV'25}
& 20.65 & 34.30 & 27.24 & 14.30 & \tabsecond{26.80} & 57.24 & 34.95 & \tabsecond{53.10} & 49.17 & 35.31 & +0.99 \\
& TAD & \pubyear{ICML'26}
& 21.67 & \tabsecond{36.45} & 27.37 & 13.00 & 23.20 & \tabbest{58.76} & \tabbest{35.41} & \tabbest{53.60} & 51.62 & 35.68 & +1.36 \\
& BiLD & \pubyear{COLING'25}
& \tabbest{25.60} & 26.60 & 27.43 & 13.02 & \tabbest{29.40} & 53.05 & 33.11 & 45.20 & 49.80 & 33.69 & -0.62 \\
& \textbf{\ALRA} & \pubyear{Ours}
& 21.33 & \tabbest{37.71} & \tabbest{28.09} & \tabsecond{19.91} & 24.20 & \tabsecond{58.65} & \tabbest{35.41} & 52.40 & \tabbest{51.85} & \tabavgbest{36.62} & \tabdeltabest{2.31} \\
\midrule
\multirow{9}{*}{\makecell[c]{Qwen1.5\\1.8B\\$\downarrow$\\500M}}
& Pre-Train w/o KD & \pubyear{--}
& 21.16 & 32.53 & 26.74 & 10.85 & 25.20 & 55.50 & 34.60 & 51.60 & 52.17 & 34.48 & Ref. \\
& Vanilla KD & \pubyear{NeurIPS'24}
& 20.90 & 35.23 & 27.26 & 15.78 & 23.40 & 57.34 & 35.16 & 52.00 & 49.01 & 35.12 & +0.64 \\
& PD & \pubyear{ACL'25}
& 21.84 & 34.30 & 26.46 & \tabsecond{19.10} & 26.00 & 58.27 & 35.01 & 51.60 & 50.59 & 35.91 & +1.42 \\
& ATKD & \pubyear{ACL'24}
& 22.10 & 37.79 & 27.93 & 17.99 & 24.60 & \tabsecond{58.65} & \tabsecond{35.88} & \tabsecond{53.40} & 50.83 & 36.57 & +2.09 \\
& RLD & \pubyear{ICCV'25}
& 21.50 & 35.31 & 27.65 & 13.24 & 26.20 & 58.49 & 34.75 & 52.90 & \tabbest{53.35} & 35.93 & +1.45 \\
& LDRLD & \pubyear{ICCV'25}
& 21.33 & 35.65 & \tabsecond{28.16} & 18.59 & 25.80 & \tabbest{59.41} & \tabbest{35.93} & 52.80 & 51.14 & 36.54 & +2.05 \\
& TAD & \pubyear{ICML'26}
& 22.35 & \tabsecond{37.96} & 27.65 & 13.60 & 24.60 & 57.83 & 35.41 & \tabsecond{53.40} & 51.30 & 36.01 & +1.53 \\
& BiLD & \pubyear{COLING'25}
& \tabbest{25.85} & 26.77 & 27.67 & 13.89 & \tabbest{29.20} & 54.08 & 33.27 & 45.10 & 49.96 & 33.98 & -0.51 \\
& \textbf{\ALRA} & \pubyear{Ours}
& \tabsecond{22.61} & \tabbest{38.26} & \tabbest{28.34} & \tabbest{20.59} & \tabsecond{26.40} & 58.60 & 35.41 & \tabbest{53.80} & \tabsecond{52.57} & \tabavgbest{37.40} & \tabdeltabest{2.91} \\
\bottomrule
\end{tabular}
\end{adjustbox}
\end{table}

Table~\ref{tab:baseline_comparison} reports zero-shot accuracy on nine
downstream benchmarks after training randomly initialized 200M- and
500M-parameter students under an approximately 1B-token budget.
\ALRA{} achieves the highest average accuracy in both settings. It reaches
36.62\% with the 200M student, exceeding TAD, the strongest competing
baseline at this size, by 0.94 percentage points. With the 500M student,
\ALRA{} obtains 37.40\%, outperforming ATKD by 0.83 percentage points.
Compared with pre-training without KD, the gains are 2.31 and 2.91
percentage points for the 200M and 500M students, respectively; the
corresponding gains over Vanilla KD are 1.49 and 2.28 points. These
improvements are also broad across tasks: \ALRA{} ranks first or second on
six of the nine benchmarks at 200M and on seven at 500M. It improves over
pre-training without KD on eight benchmarks at 200M and over both
pre-training without KD and Vanilla KD on all nine benchmarks at 500M.

Without KD, the student is trained only with the causal language-modeling
loss. Vanilla KD additionally matches the teacher distribution over the
full vocabulary, providing information about alternative tokens, but its
distillation term does not distinguish between local and rest regions.
PD improves over Vanilla KD at both model sizes by applying
top-\(p\)-\(k\) truncation and renormalizing the teacher distribution over
the retained high-probability tokens \citep{peng2025pre}. However,
its token selection is determined only by the teacher distribution and
fixed truncation parameters, whereas \ALRA{} begins from a
student-proposed candidate set.

ATKD adapts the distillation loss across prediction positions using the
teacher's uncertainty coefficient. It omits target-oriented KD for easy
positions and applies both target- and diversity-oriented KD to hard
positions \citep{zhong2024revisiting}. TAD instead decomposes the KL loss
into top-\(K\) and tail components and normalizes the tail contribution
using the average teacher tail mass over each sequence
\citep{dasguptadon}. Both methods adapt the distillation objective,
but TAD uses the same \(K\) at every position. In contrast, \ALRA{} forms a
student-proposed candidate set, inserts the teacher top-1 token when
needed, and uses the teacher probabilities within this anchored set to
determine a position-specific budget \(d_u\) and select the final local
tokens. The teacher top-1 token serves only as an anchor, not as a
ground-truth label. Adaptive Local Divergence then retains all remaining
tokens in the rest region and aligns both the probability mass and the
conditional distribution within the two regions.

RLD improves over Vanilla KD by 0.34 and 0.81 percentage points for the
200M and 500M students, respectively. It uses label information to align
the student's true-class confidence with the teacher's maximum confidence
and masks classes whose teacher logits are at least as large as the
true-class logit before aligning the remaining classes
\citep{sun2025knowledge}. \ALRA{} exceeds RLD by 1.15 and 1.47 percentage
points. The two methods also differ in their use of labels: RLD was
designed around a true class in image classification, whereas \ALRA{} does
not use the target token when forming its local token set.

LDRLD \citep{xu2025local} is the closest relational baseline and
motivates the pairwise component of \ALRA. It selects the top-\(d\)
student logits and transfers relations among the selected classes, but
the recursion depth \(d\) is manually chosen and fixed throughout a
training run. Its local selection also depends entirely on the student
ranking, which may be unreliable early in training for a randomly
initialized student. \ALRA{} instead uses a position-specific budget
\(d_u\) and adds the teacher top-1 token before forming the final local
set. The teacher probabilities over this anchored candidate set then
refine the student proposal and retain the more strongly preferred
candidates for local supervision. The methods also differ in pair
weighting: LDRLD derives weights from rank difference and rank sum,
whereas SWPRA uses the student's current full-vocabulary probabilities,
giving more weight to high-mass pairs with small probability gaps.
\ALRA{} exceeds LDRLD by 1.31 and 0.86 percentage points for the 200M and
500M students, respectively.

BiLD achieves the best results on ARC-Challenge and OpenBookQA, but its
average remains below pre-training without KD at both student sizes. It
constructs pairwise logit differences from teacher-led and student-led
top-\(k\) sets and was originally developed for task-specific
distillation of existing language-model checkpoints
\citep{li2025bild}. This differs from our setting, where the students are
randomly initialized and distilled during pre-training. The student-led
top-\(k\) branch may therefore be less reliable early in training, which
may contribute to the lower average, although
Table~\ref{tab:baseline_comparison} does not isolate the exact cause.
BiLD also restricts alignment to selected top-\(k\) logits, whereas
\ALRA{} retains the remaining vocabulary in the rest region and continues
to align its conditional distribution.

\subsection{Analysis of Adaptive Local Budget Allocation}
\label{subsec:adaptive_budget_ablation}

\begin{table}[!tbp]
\caption{
Average validation accuracy (\%) after approximately 550M training
tokens. In the Fixed-Budget settings, the same
\(d\) is used at every prediction position, and the local set contains
the \(d\) vocabulary tokens assigned the highest probabilities by the
student. These settings use ALD and PRA-U, without ground-truth
grounding or force-including the teacher top-1 token. \ALRA{} uses the
complete proposed pipeline with \(d_{\min}=3\) and \(d_{\max}=25\),
including teacher top-1 anchoring, the adaptive local budget, ALD, and
SWPRA. ``Avg.'' is the arithmetic mean over the ARC-Easy, HellaSwag,
and PIQA validation sets. \(\Delta\mathrm{Avg.}\uparrow\) is the absolute
difference from the \(d=3\) Fixed-Budget setting under the same student
size. A/G/T denote adaptive
local budget, ground-truth anchoring, and teacher top-1 anchoring,
respectively.
}
\label{tab:validation_fixed_budget_ablation}
\centering
\scriptsize
\setlength{\tabcolsep}{5.0pt}
\renewcommand{\arraystretch}{1.14}
\begin{adjustbox}{max width=\textwidth}
\begin{tabular}{@{}>{\centering\arraybackslash}m{1.65cm}
                    >{\centering\arraybackslash}m{2.85cm}
                    >{\centering\arraybackslash}m{1.45cm}
                    ccc
                    >{\centering\arraybackslash}m{1.45cm}
                    >{\centering\arraybackslash}m{0.80cm}
                    >{\centering\arraybackslash}m{1.05cm}@{}}
\toprule
\multirow{2}{*}{\makecell[c]{\textbf{Teacher/}\\\textbf{Student}}} &
\multirow{2}{*}{\textbf{Objective}} &
\multirow{2}{*}{\makecell[c]{\textbf{Local}\\\textbf{budget}}} &
\multicolumn{4}{c}{\textbf{Components}} &
\multirow{2}{*}{\textbf{Avg.}} &
\multirow{2}{*}{$\boldsymbol{\Delta}_{\mathrm{Avg.}}\!\uparrow$} \\
\cmidrule(lr){4-7}
& & & \textbf{A} & \textbf{G} & \textbf{T} & \textbf{Pairwise loss} & & \\
\midrule
\multirow{7}{*}{\makecell[c]{Qwen1.5\\1.8B\\$\downarrow$\\200M}}
& \multirow{6}{*}{\makecell[c]{\FixedBudget}} & $d=3$  & \xmark & \xmark & \xmark & \PRAU & 37.91 & Ref. \\
& & $d=7$  & \xmark & \xmark & \xmark & \PRAU & 38.54 & +0.63 \\
& & $d=9$  & \xmark & \xmark & \xmark & \PRAU & 38.99 & +1.08 \\
& & $d=15$ & \xmark & \xmark & \xmark & \PRAU & 39.39 & +1.48 \\
& & $d=19$ & \xmark & \xmark & \xmark & \PRAU & 39.07 & +1.16 \\
& & $d=25$ & \xmark & \xmark & \xmark & \PRAU & 39.20 & +1.29 \\
& \textbf{\ALRA} & \textbf{$d_u\in[3,25]$} & \textbf{\cmark} & \textbf{\xmark} & \textbf{\cmark} & SWPRA & \tabavgbest{39.89} & \tabdeltabest{1.98} \\
\midrule
\multirow{7}{*}{\makecell[c]{Qwen1.5\\1.8B\\$\downarrow$\\500M}}
& \multirow{6}{*}{\makecell[c]{\FixedBudget}} & $d=3$  & \xmark & \xmark & \xmark & \PRAU & 39.27 & Ref. \\
& & $d=7$  & \xmark & \xmark & \xmark & \PRAU & 39.43 & +0.16 \\
& & $d=9$  & \xmark & \xmark & \xmark & \PRAU & 40.67 & +1.40 \\
& & $d=15$ & \xmark & \xmark & \xmark & \PRAU & 40.67 & +1.40 \\
& & $d=19$ & \xmark & \xmark & \xmark & \PRAU & 40.73 & +1.46 \\
& & $d=25$ & \xmark & \xmark & \xmark & \PRAU & 40.69 & +1.42 \\
& \textbf{\ALRA} & \textbf{$d_u\in[3,25]$} & \textbf{\cmark} & \textbf{\xmark} & \textbf{\cmark} & SWPRA & \tabavgbest{41.09} & \tabdeltabest{1.82} \\
\bottomrule
\end{tabular}
\end{adjustbox}
\end{table}

Table~\ref{tab:validation_fixed_budget_ablation} compares six fixed
local budgets with the complete \ALRA{} configuration. In each
Fixed-Budget setting, the same \(d\) is used at every prediction
position. The local set is taken directly as the \(d\) vocabulary tokens
assigned the highest probabilities by the student, while all remaining
tokens form the rest region. ALD is computed over this local--rest
partition, and PRA-U gives every token pair in the local set the same
normalized weight. The teacher top-1 token is not force-included, so it
belongs to the local set only when it is already ranked among these
\(d\) tokens by the student.

Increasing \(d\) improves validation accuracy at first, but the trend is
not monotonic and the best value differs between the two student sizes.
For the 200M student, the best fixed-budget result is 39.39 at \(d=15\);
increasing \(d\) to 19 or 25 lowers the accuracy. For the 500M student,
the best result instead occurs at \(d=19\), with 40.73, while \(d=25\)
gives no further improvement. A fixed-budget method therefore requires
\(d\) to be chosen as a hyperparameter, and Table~\ref{tab:validation_fixed_budget_ablation}
does not indicate one fixed value that works best for both students.

The fixed construction also lets the student ranking fully determine
which tokens enter the local set. This matters especially early in
training, when the randomly initialized student may not yet rank
teacher-preferred tokens highly. The teacher top-1 token can therefore
remain outside the local set, so the local set is not guaranteed to contain the teacher top-1 token. Such a token is still included in
the rest region and supervised by ALD, but it does not take part in the
local conditional distribution or the local pairwise relations. In
other words, the teacher still provides the distillation targets, but it
does not guide which tokens are placed in the local set. PRA-U also
weights all selected pairs equally after normalization, without
distinguishing them by the student's probability mass or probability
gap.

\ALRA{} changes this construction while keeping
\(3 \leq d_u \leq 25\), bounded by the smallest and largest fixed budgets
evaluated in the table. It includes the teacher top-1 token in the
candidate set, chooses \(d_u\) separately for each prediction position,
uses teacher probabilities to select the final local tokens, and applies
SWPRA to weight the local pairs. The complete configuration reaches
39.89 for the 200M student and 41.09 for the 500M student, exceeding the
best fixed-budget results by 0.50 and 0.36 percentage points,
respectively. These results show that the complete \ALRA{} configuration
achieves higher validation accuracy than every tested fixed-budget
setting for both student sizes. Since the \ALRA{} row also changes the
candidate-set construction and pair weighting, the next analysis examines teacher top-1 and ground-truth anchoring,
followed by a separate comparison of the pair-weighting schemes.

\begin{figure}
    \centering
    \includegraphics[width=0.88\textwidth, keepaspectratio]
    {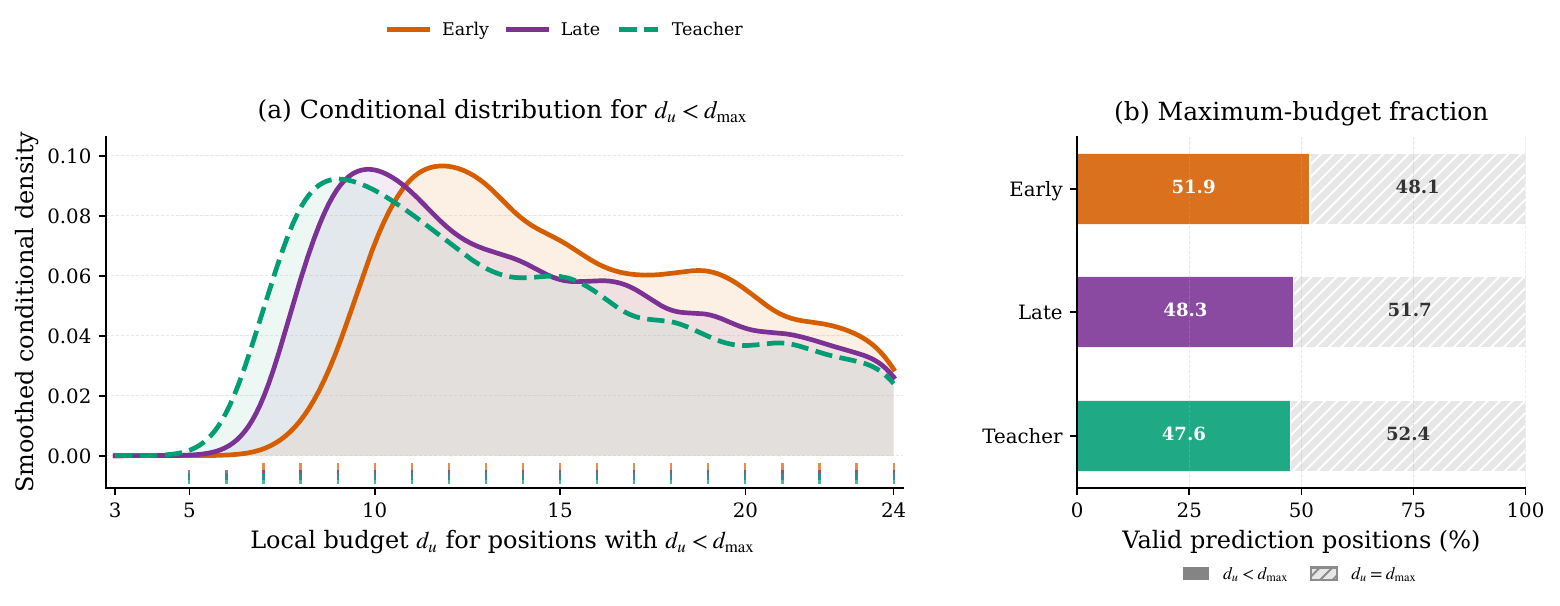}
    \caption{
    Distribution of the position-specific local budget \(d_u\), with
    \(d_{\min}=3\) and \(d_{\max}=25\), during a 1B-token training run.
    Early and Late are sampled near the beginning and end of student
    training, respectively. Teacher refers to the frozen teacher and is shown only as a reference,
    not as a stage of student training. For this Teacher reference, the
    \(d_{\max}\) vocabulary tokens with the highest probabilities under the
    teacher form the candidate set at each sampled prediction position.
    The candidate-conditioned effective support and its batch average are
    then computed, and the same adaptive-budget rule is used to obtain
    \(d_u\).
    (a) Smoothed conditional density of \(d_u\) for sampled valid prediction
    positions with \(d_u<d_{\max}\); positions with \(d_u=d_{\max}\) are
    excluded from this panel.
    (b) Fractions of sampled valid prediction positions with
    \(d_u<d_{\max}\) and \(d_u=d_{\max}\).
    }
    \label{fig:budget_dynamics}
\end{figure}

Figure~\ref{fig:budget_dynamics} shows how the position-specific budget
\(d_u\) is assigned at different stages of training. Panel~(a) includes
only positions with \(d_u<d_{\max}\) and shows how these budget values
are distributed. Panel~(b) uses all sampled positions and reports the
fractions assigned a budget below \(d_{\max}\) or exactly
\(d_{\max}\).

\emph{Early stage (orange).}
At the beginning of training, the student is randomly initialized, so
tokens ranked highly by the student can receive little probability from
the teacher. After the teacher top-1 token is inserted, the teacher
distribution within the anchored candidate set can become strongly
concentrated around this token, giving a small
\(E_u^{\mathrm{loc}}\). If this occurs at many positions in the same
batch, the batch average \(\bar{E}_{\Tbatch}^{\mathrm{loc}}\) is also
small. From Eq.~\eqref{eq:adaptive_local_budget}, the ratio
\(E_u^{\mathrm{loc}}/\bar{E}_{\Tbatch}^{\mathrm{loc}}\) can therefore
remain close to one, leading to large \(d_u\) values and, for some
positions, \(d_{\max}\). This is consistent with the Early curve in
panel~(a), which is shifted toward larger non-maximum budgets.

A larger \(d_u\) allows more candidates with high teacher probability
to remain in the final local set. At this early stage, the student
therefore receives local and pairwise supervision over more token
alternatives, rather than having the local information narrowed too
strongly while its own ranking is still unreliable.

\emph{Late stage (purple).}
As training progresses, the student proposals can contain more tokens
that also receive meaningful probability from the teacher. The teacher
distribution within the candidate set can then take different forms
across prediction positions. At positions where several candidate
tokens receive meaningful teacher probability, the distribution is more
diffuse and \(E_u^{\mathrm{loc}}\) is larger. If it is also large
relative to the current batch average, Eq.~\eqref{eq:adaptive_local_budget}
assigns a larger \(d_u\), so more competing alternatives remain in the
final local set.

At other positions, the teacher probability is concentrated on fewer
candidates. When the corresponding effective support is smaller
relative to the batch average, a smaller \(d_u\) is assigned. The final
teacher ranking then keeps a smaller set of candidates with stronger
teacher preference. In this way, the local-set size can change from one
prediction position to another according to how the teacher evaluates
the candidates proposed by the current student.

Panel~(a) shows this change among positions with
\(d_u<d_{\max}\). The Late curve shifts toward smaller budgets than the
Early curve and peaks around \(d_u\approx9\)--\(10\). This does not mean
that \(d_u\) becomes smaller at every position, because positions with
\(d_u=d_{\max}\) are excluded from panel~(a). Panel~(b) shows that the
fraction assigned \(d_{\max}\) instead increases slightly from 48.1\%
at Early to 51.7\% at Late. Thus, later in training, smaller local sets
become more common among the non-maximum positions, while many other
positions still receive the maximum budget.

\emph{Teacher reference (green dashed).}
Teacher denotes the frozen teacher and is not a stage of student
training. At each sampled prediction position, the \(d_{\max}\)
vocabulary tokens with the highest teacher probabilities form its
candidate set. Since the teacher top-1 token is already contained in
this set, no additional anchor changes the candidate set. The same
candidate-conditioned effective-support, batch-average, and adaptive
budget calculations in Eqs.~\eqref{eq:candidate_conditioned_teacher_support}--%
\eqref{eq:adaptive_local_budget} are then used to obtain \(d_u\).

Among positions with \(d_u<d_{\max}\), the Teacher curve is shifted
toward smaller budgets than the Early curve, while the Late curve is
closer to the Teacher reference. This means that the shape of the
student's non-maximum budget distribution becomes more similar to the
Teacher reference later in training. It does not imply that the student
has matched the teacher's downstream performance.

Panel~(b) shows that 48.1\%, 51.7\%, and 52.4\% of the sampled positions
receive \(d_{\max}\) for Early, Late, and Teacher, respectively. The
remaining positions receive smaller, position-specific budgets. Thus,
the adaptive rule uses both maximum and non-maximum budgets instead of
assigning \(d_{\max}\) to every position.

\FloatBarrier

\subsection{Effect of Teacher Top-1 and Ground-Truth Anchoring}
\label{subsec:grounding_ablation}

\begin{table}[!tbp]
\caption{
Average validation accuracy (\%) after approximately 550M training
tokens. All variants use the same adaptive-budget rule, ALD, and SWPRA;
they differ only in how the candidate set is anchored: no anchor,
ground-truth only, teacher top-1 only, or both.
``Avg.'' is the arithmetic mean over the ARC-Easy, HellaSwag, and PIQA
validation sets. \(\Delta_{\mathrm{Avg.}}\!\uparrow\) is the absolute
difference from \ALRAN{} under the same student size. A/G/T denote
adaptive local budget, ground-truth anchoring, and teacher top-1
anchoring, respectively.
}
\label{tab:validation_grounding_ablation}
\centering
\scriptsize
\setlength{\tabcolsep}{5.2pt}
\renewcommand{\arraystretch}{1.14}
\begin{adjustbox}{max width=0.86\textwidth}
\begin{tabular}{@{}>{\centering\arraybackslash}m{1.75cm}
                    >{\centering\arraybackslash}m{1.65cm}
                    ccc
                    >{\centering\arraybackslash}m{1.55cm}
                    >{\centering\arraybackslash}m{0.85cm}
                    >{\centering\arraybackslash}m{1.10cm}@{}}
\toprule
\multirow{2}{*}{\makecell[c]{\textbf{Teacher/}\\\textbf{Student}}} &
\multirow{2}{*}{\textbf{Method}} &
\multicolumn{4}{c}{\textbf{Components}} &
\multirow{2}{*}{\textbf{Avg.}} &
\multirow{2}{*}{$\boldsymbol{\Delta}_{\mathrm{Avg.}}\!\uparrow$} \\
\cmidrule(lr){3-6}
& & \textbf{A} & \textbf{G} & \textbf{T} & \textbf{Pairwise loss} & & \\
\midrule
\multirow{4}{*}{\makecell[c]{Qwen1.5\\1.8B\\$\downarrow$\\200M}}
& \ALRAN  & \cmark & \xmark & \xmark & \SWPRA & 38.86 & Ref. \\
& \ALRANG & \cmark & \cmark & \xmark & \SWPRA & 38.87 & +0.01 \\
& \textbf{\ALRA} & \textbf{\cmark} & \textbf{\xmark} &
  \textbf{\cmark} & \SWPRA &
  \tabavgbest{39.89} & \tabdeltabest{1.03} \\
& \ALRAG  & \cmark & \cmark & \cmark & \SWPRA & 39.21 & +0.35 \\
\midrule
\multirow{4}{*}{\makecell[c]{Qwen1.5\\1.8B\\$\downarrow$\\500M}}
& \ALRAN  & \cmark & \xmark & \xmark & \SWPRA & 40.07 & Ref. \\
& \ALRANG & \cmark & \cmark & \xmark & \SWPRA & 40.74 & +0.67 \\
& \textbf{\ALRA} & \textbf{\cmark} & \textbf{\xmark} &
  \textbf{\cmark} & \SWPRA &
  \tabavgbest{41.09} & \tabdeltabest{1.02} \\
& \ALRAG  & \cmark & \cmark & \cmark & \SWPRA & 40.52 & +0.45 \\
\bottomrule
\end{tabular}
\end{adjustbox}
\end{table}

Table~\ref{tab:validation_grounding_ablation} examines how candidate-set
anchoring affects the adaptive local set. The four variants use the same
adaptive-budget rule, ALD, and SWPRA. For \ALRAN{}, the initial candidate
set is exactly the student's top-\(d_{\max}\) proposal, with no added
anchor. \ALRANG{} adds the ground-truth next token \(y_u\), \ALRA{} adds
only the teacher top-1 token as defined in
Eq.~\eqref{eq:teacher_anchored_candidate_set}, and \ALRAG{} adds both.
The candidate set always contains \(d_{\max}\) tokens; any added anchor
that is not already present replaces a lower-ranked student proposal.

Without either anchor, \ALRAN{} gives the lowest validation average for
both students, with 38.86 at 200M and 40.07 at 500M. The teacher is
still used in this variant: its probabilities are used to compute the
candidate-conditioned effective support, determine \(d_u\), and select
the final local tokens. The limitation is that the initial candidate set
comes entirely from the student's ranking. Early in training, this
ranking can be unreliable, so the teacher top-1 token may be absent from
the candidate set. If it is absent at this stage, the later teacher
ranking cannot select it because it can only rank tokens already present
in the candidate set.

Teacher top-1 anchoring gives the largest improvement. Compared with
\ALRAN{}, \ALRA{} increases the validation average by 1.03 points for
the 200M student and 1.02 points for the 500M student, giving the best
result in both settings. The anchor guarantees that the token with the
highest teacher probability is already present when the effective
support and \(d_u\) are computed. The final local set is then selected
by teacher probability, so the teacher top-1 token is also retained in
that set. Most candidates still come from the student proposal, while
the local set is guaranteed to contain the token most preferred by the
teacher.

\begin{figure}
    \centering
    \includegraphics[width=\textwidth, keepaspectratio]
    {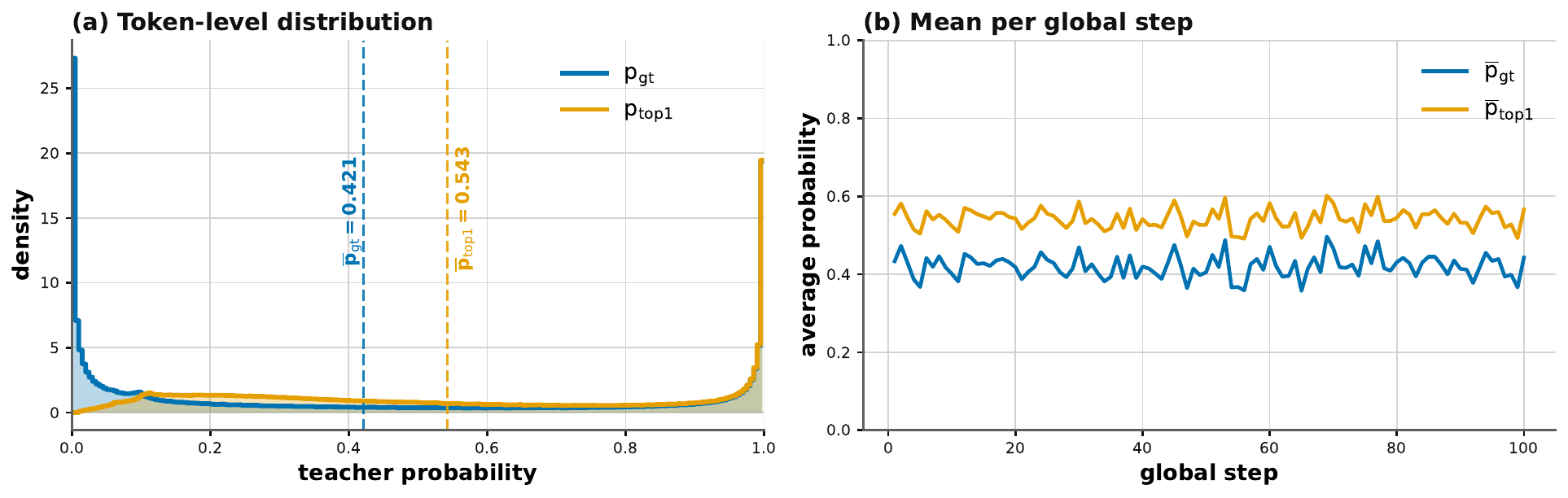}
    \caption{
    Teacher probability assigned to the ground-truth next token
    (\(p_{\mathrm{gt}}\)) and the teacher top-1 token
    (\(p_{\mathrm{top1}}\)), measured with the frozen teacher over
    8{,}000 sampled sequences (\(\approx\)3.9M tokens). These probabilities
    depend only on the frozen teacher and the sampled prediction positions,
    not on the student.
    (a) Token-level probability density. The density of
    \(p_{\mathrm{gt}}\) is high near zero, showing that the teacher assigns
    low probability to the ground-truth token at many sampled positions.
    (b) Mean probability at each sampled global step.
    \(\bar{p}_{\mathrm{top1}}\) remains above
    \(\bar{p}_{\mathrm{gt}}\) across the sampled steps.
    Across all sampled positions,
    \(\bar{p}_{\mathrm{gt}}=0.421\) and
    \(\bar{p}_{\mathrm{top1}}=0.543\), a difference of \(0.122\).
    The teacher top-1 and ground-truth tokens coincide at 53\% of the
    sampled positions.
    }
    \label{fig:teacher_gt_vs_top1}
\end{figure}

Ground-truth anchoring gives a different result. \ALRANG{} changes the
200M validation average only slightly, from 38.86 to 38.87, but improves
the 500M result from 40.07 to 40.74. Unlike the teacher top-1 token, the
ground-truth token is guaranteed only to enter the candidate set. It is
not guaranteed to remain in the final local set, because the final
selection is still based on teacher probability.
Figure~\ref{fig:teacher_gt_vs_top1} helps explain why this difference
matters.

Since \(p_{\mathrm{top1}}\geq p_{\mathrm{gt}}\) by definition, the main
point of Fig.~\ref{fig:teacher_gt_vs_top1} is not this inequality
itself, but how often the two tokens differ and how much probability the
teacher assigns to the ground-truth token. The two tokens coincide at
53\% of the sampled positions. At many other positions,
\(p_{\mathrm{gt}}\) is close to zero, as shown by the high density near
zero in panel~(a). Panel~(b) shows the same pattern across sampled
global steps: although the data change from step to step, the mean
teacher top-1 probability remains above the mean ground-truth
probability.

This directly affects the candidate-set construction. Force-including a
ground-truth token with low teacher probability changes the
candidate-conditioned distribution used to compute
\(E_u^{\mathrm{loc}}\), and can therefore also change the resulting
\(d_u\). However, because the final local set is selected by teacher
probability, the ground-truth token can still be removed when its teacher
probability is low. Ground-truth anchoring therefore guarantees that the
observed next token is considered when the candidate set is formed, but
does not guarantee that it will be used in the final local supervision.
This is consistent with the different gains of \ALRANG{} for the two
student sizes.

Adding both anchors does not improve over teacher top-1 anchoring
alone. \ALRAG{} reaches 39.21 and 40.52, which are 0.68 and 0.57 points
below \ALRA{} for the 200M and 500M students, respectively. When the
ground-truth and teacher top-1 tokens are the same, adding both does not
introduce another distinct token.  When they differ, both tokens must be included in the candidate set of
\(d_{\max}\) tokens, leaving fewer positions for the original student
proposals. The additional ground-truth token can also change the
candidate-conditioned effective support even when it receives low
teacher probability and is not retained in the final local set. These changes in the candidate set can explain why the two anchors do
not necessarily provide additive gains.

The comparison therefore supports teacher top-1 anchoring as the choice used in the final \ALRA{} configuration. It gives the highest validation average for both student sizes among the four variants in Table~\ref{tab:validation_grounding_ablation}. Ground-truth anchoring does not reach the same performance, either when used alone or when added together with the teacher top-1 anchor. Based on these results, \ALRA{} uses only the teacher top-1 token as the candidate-set anchor.


\subsection{Effect of Pairwise Weighting}
\label{subsec:swpra_analysis}

\begin{table}[!tbp]
\caption{
Held-out zero-shot average accuracy (\%) after approximately 1B training
tokens. All variants use the same teacher-anchored adaptive local token
set and ALD and differ only in how the pairwise terms are weighted.
\ALRAR{} uses PRA with unit pair weights, \ALRAU{} uses PRA-U with
normalized uniform weights, and \ALRA{} uses SWPRA with the normalized
student-dependent weights in Eq.~\eqref{eq:normalized_pair_weight}.``Avg.'' is the
arithmetic mean over the same nine held-out benchmarks used in
Table~\ref{tab:baseline_comparison}. \(\Delta_{\mathrm{Avg.}}\!\uparrow\)
is the absolute difference from \ALRAR{} under the same student size.
}
\label{tab:pair_weighting_ablation}
\centering
\scriptsize
\setlength{\tabcolsep}{6.2pt}
\renewcommand{\arraystretch}{1.14}
\begin{adjustbox}{max width=0.78\textwidth}
\begin{tabular}{@{}llcccccc@{}}
\toprule
\multirow{2}{*}{\textbf{Teacher/Student}} &
\multirow{2}{*}{\textbf{Method}} &
\multicolumn{3}{c}{\textbf{Pairwise loss}} &
\multirow{2}{*}{\textbf{Avg.}} &
\multirow{2}{*}{$\boldsymbol{\Delta}_{\mathrm{Avg.}}\!\uparrow$} \\
\cmidrule(lr){3-5}
& & \textbf{PRA} & \textbf{PRA-U} & \textbf{SWPRA} & & \\
\midrule
\multirow{3}{*}{\makecell[c]{Qwen1.5-1.8B\\$\downarrow$\\200M}}
& \ALRAR & \cmark & \xmark & \xmark & 34.91 & Ref. \\
& \ALRAU & \xmark & \cmark & \xmark & 36.56 & +1.65 \\
& \textbf{\ALRA} & \textbf{\xmark} & \textbf{\xmark} &
  \textbf{\cmark} & \tabavgbest{36.62} & \tabdeltabest{1.71} \\
\midrule
\multirow{3}{*}{\makecell[c]{Qwen1.5-1.8B\\$\downarrow$\\500M}}
& \ALRAR & \cmark & \xmark & \xmark & 35.77 & Ref. \\
& \ALRAU & \xmark & \cmark & \xmark & 37.09 & +1.32 \\
& \textbf{\ALRA} & \textbf{\xmark} & \textbf{\xmark} &
  \textbf{\cmark} & \tabavgbest{37.40} & \tabdeltabest{1.63} \\
\bottomrule
\end{tabular}
\end{adjustbox}
\end{table}

Table~\ref{tab:pair_weighting_ablation} isolates the effect of pairwise
weighting. All three variants use the same local token pairs
\(\Pset_u\) from Eq.~\eqref{eq:pair_set} and the same teacher-to-student
pairwise KL in Eq.~\eqref{eq:general_pairwise_loss}; only the pair
weights are changed.

PRA assigns weight one to every pair. Since
\(|\Pset_u|=\binom{d_u}{2}\), a larger \(d_u\) produces more pairwise
terms and therefore a larger summed pairwise loss simply because more
pairs are present. PRA-U removes this direct dependence by assigning
each pair weight \(1/|\Pset_u|\) in Eq.~\eqref{eq:pra_variants}. The
pairwise term then becomes an average over the local pairs rather than
an unnormalized sum. This gives the student a more comparable pairwise
signal across prediction positions with different \(d_u\). Consistent
with this change, PRA-U improves the average from 34.91 to 36.56 for
the 200M student and from 35.77 to 37.09 for the 500M student.

SWPRA keeps the same normalization but changes how the total pair weight
is distributed. From Eq.~\eqref{eq:unnormalized_pair_score}, a pair
receives a larger score when the two tokens carry more student
probability mass and have a smaller student probability gap. Intuitively,
these pairs correspond to plausible local alternatives that the student
still assigns similar probabilities to. Giving them more weight focuses
the pairwise supervision on comparisons for which the student's relative
preference is still weak, while reducing the weight of low-probability
pairs or pairs that the student already separates strongly. The scores are normalized in
Eq.~\eqref{eq:normalized_pair_weight}, so SWPRA redistributes the
pairwise emphasis across the local pairs according to the current
student distribution, rather than weighting all pairs uniformly.

Compared with PRA-U, SWPRA further increases the average from 36.56 to
36.62 for the 200M student and from 37.09 to 37.40 for the 500M student.
The gain is small at 200M but larger at 500M, so the main benefit in
Table~\ref{tab:pair_weighting_ablation} comes from normalizing the
pairwise term, while student-dependent weighting provides an additional
improvement. Since SWPRA gives the highest average for both student
sizes, it is used as the pairwise term in the final \ALRA{} objective
in Eq.~\eqref{eq:overall_alra_loss}.

\FloatBarrier

\section{Conclusion}
\label{sec:conclusion}

We introduced Adaptive Local Relational Alignment (ALRA) for
logit-based pre-training distillation of autoregressive language models.
ALRA combines student-proposed tokens with teacher guidance to construct
a position-specific local token set. Adaptive Local Divergence (ALD)
uses the resulting local--rest partition to align probability mass and
the conditional distributions within both regions, while
Student-Weighted Pairwise Relational Alignment (SWPRA) further aligns
relative preferences among local token pairs, with greater emphasis on
pairs with high student probability mass and small probability gaps.
Across nine zero-shot benchmarks, ALRA achieves average accuracies of
36.62\% and 37.40\% for randomly initialized 200M- and 500M-parameter
students distilled from a frozen Qwen1.5-1.8B teacher, exceeding the
strongest competing distillation baseline at each student size by 0.94
and 0.83 percentage points, respectively. The controlled studies also
show the roles of candidate-set anchoring and pairwise weighting and
compare the complete ALRA configuration with fixed local budgets. The
current formulation of ALRA is limited to white-box logit distillation in which
the teacher and student share the same output vocabulary. Extending the
framework to settings with different tokenizers or more limited access
to teacher outputs would require a different way to define and align
the local token space. In addition, this work focuses on token-level pre-training distillation for autoregressive language models. Combining ALRA with sequence-level distillation or other stages of language-model training remains an open direction for future work.

\FloatBarrier


\printcredits

\bibliographystyle{cas-model2-names}
\bibliography{cas-refs}

@article{hinton2015distilling,
  title={Distilling the knowledge in a neural network},
  author={Hinton, Geoffrey and Vinyals, Oriol and Dean, Jeff},
  journal={arXiv preprint arXiv:1503.02531},
  year={2015}
}

@article{yang2025survey,
  title={Survey on knowledge distillation for large language models: methods, evaluation, and application},
  author={Yang, Chuanpeng and Zhu, Yao and Lu, Wang and Wang, Yidong and Chen, Qian and Gao, Chenlong and Yan, Bingjie and Chen, Yiqiang},
  journal={ACM Transactions on Intelligent Systems and Technology},
  volume={16},
  number={6},
  pages={1--27},
  year={2025},
  publisher={ACM New York, NY}
}

@inproceedings{zhong2024revisiting,
  title={Revisiting knowledge distillation for autoregressive language models},
  author={Zhong, Qihuang and Ding, Liang and Shen, Li and Liu, Juhua and Du, Bo and Tao, Dacheng},
  booktitle={Proceedings of the 62nd Annual Meeting of the Association for Computational Linguistics (Volume 1: Long Papers)},
  pages={10900--10913},
  year={2024}
}

@article{fang2025knowledge,
  title={Knowledge distillation and dataset distillation of large language models: Emerging trends, challenges, and future directions},
  author={Fang, Luyang and Yu, Xiaowei and Cai, Jiazhang and Chen, Yongkai and Wu, Shushan and Liu, Zhengliang and Yang, Zhenyuan and Lu, Haoran and Gong, Xilin and Liu, Yufang and others},
  journal={Artificial Intelligence Review},
  volume={59},
  number={1},
  pages={17},
  year={2025},
  publisher={Springer}
}

@article{sanh2019distilbert,
  title={DistilBERT, a distilled version of BERT: smaller, faster, cheaper and lighter},
  author={Sanh, Victor and Debut, Lysandre and Chaumond, Julien and Wolf, Thomas},
  journal={arXiv preprint arXiv:1910.01108},
  year={2019}
}

@inproceedings{jiao2020tinybert,
  title={Tinybert: Distilling bert for natural language understanding},
  author={Jiao, Xiaoqi and Yin, Yichun and Shang, Lifeng and Jiang, Xin and Chen, Xiao and Li, Linlin and Wang, Fang and Liu, Qun},
  booktitle={Findings of the association for computational linguistics: EMNLP 2020},
  pages={4163--4174},
  year={2020}
}

@article{wang2020minilm,
  title={Minilm: Deep self-attention distillation for task-agnostic compression of pre-trained transformers},
  author={Wang, Wenhui and Wei, Furu and Dong, Li and Bao, Hangbo and Yang, Nan and Zhou, Ming},
  journal={Advances in neural information processing systems},
  volume={33},
  pages={5776--5788},
  year={2020}
}

@inproceedings{sun2020mobilebert,
  title={Mobilebert: a compact task-agnostic bert for resource-limited devices},
  author={Sun, Zhiqing and Yu, Hongkun and Song, Xiaodan and Liu, Renjie and Yang, Yiming and Zhou, Denny},
  booktitle={Proceedings of the 58th annual meeting of the association for computational linguistics},
  pages={2158--2170},
  year={2020}
}

@inproceedings{dasguptadon,
  title={Don't Ignore the Tail: Decoupled Distillation Produces Top Maths Students on an Academic Budget},
  author={Dasgupta, Sayantan and Cohn, Trevor and Baldwin, Timothy},
  booktitle={Forty-third International Conference on Machine Learning},
  year={2026}
}

@article{muralidharan2024compact,
  title={Compact language models via pruning and knowledge distillation},
  author={Muralidharan, Saurav and Turuvekere Sreenivas, Sharath and Joshi, Raviraj and Chochowski, Marcin and Patwary, Mostofa and Shoeybi, Mohammad and Catanzaro, Bryan and Kautz, Jan and Molchanov, Pavlo},
  journal={Advances in Neural Information Processing Systems},
  volume={37},
  pages={41076--41102},
  year={2024}
}

@article{tang2019distilling,
  title={Distilling task-specific knowledge from bert into simple neural networks},
  author={Tang, Raphael and Lu, Yao and Liu, Linqing and Mou, Lili and Vechtomova, Olga and Lin, Jimmy},
  journal={arXiv preprint arXiv:1903.12136},
  year={2019}
}

@inproceedings{peng2025pre,
  title={Pre-training distillation for large language models: A design space exploration},
  author={Peng, Hao and Lv, Xin and Bai, Yushi and Yao, Zijun and Zhang, Jiajie and Hou, Lei and Li, Juanzi},
  booktitle={Proceedings of the 63rd Annual Meeting of the Association for Computational Linguistics (Volume 1: Long Papers)},
  pages={3603--3618},
  year={2025}
}

@inproceedings{park2019relational,
  title={Relational knowledge distillation},
  author={Park, Wonpyo and Kim, Dongju and Lu, Yan and Cho, Minsu},
  booktitle={2019 IEEE/CVF Conference on Computer Vision and Pattern Recognition (CVPR)},
  pages={3962--3971},
  year={2019},
  organization={IEEE}
}

@inproceedings{zhang2023reaugkd,
  title={Reaugkd: Retrieval-augmented knowledge distillation for pre-trained language models},
  author={Zhang, Jianyi and Muhamed, Aashiq and Anantharaman, Aditya and Wang, Guoyin and Chen, Changyou and Zhong, Kai and Cui, Qingjun and Xu, Yi and Zeng, Belinda and Chilimbi, Trishul and others},
  booktitle={Proceedings of the 61st Annual Meeting of the Association for Computational Linguistics (Volume 2: Short Papers)},
  pages={1128--1136},
  year={2023}
}

@inproceedings{sun2025knowledge,
  title={Knowledge distillation with refined logits},
  author={Sun, Wujie and Chen, Defang and Lyu, Siwei and Chen, Genlang and Chen, Chun and Wang, Can},
  booktitle={2025 IEEE/CVF International Conference on Computer Vision (ICCV)},
  pages={1110--1119},
  year={2025},
  organization={IEEE}
}

@inproceedings{xu2025local,
  title={Local dense logit relations for enhanced knowledge distillation},
  author={Xu, Liuchi and Liu, Kang and Liu, Jinshuai and Wang, Lu and Xu, Lisheng and Cheng, Jun},
  booktitle={2025 IEEE/CVF International Conference on Computer Vision (ICCV)},
  pages={4539--4549},
  year={2025},
  organization={IEEE}
}

@inproceedings{he2025kd,
  title={{DA-KD}: Difficulty-Aware Knowledge Distillation for Efficient Large Language Models},
  author={He, Changyi and Ding, Yifu and Guo, Jinyang and Gong, Ruihao and Qin, Haotong and Liu, Xianglong},
  booktitle={Proceedings of the 42nd International Conference on Machine Learning},
  volume={267},
  series={Proceedings of Machine Learning Research},
  pages={22379--22391},
  year={2025},
  publisher={PMLR}
}

@article{tavor2026rethinking,
  title={Rethinking selective knowledge distillation},
  author={Tavor, Almog and Ebenspanger, Itay and Cnaan, Neil and Geva, Mor},
  journal={arXiv preprint arXiv:2602.01395},
  year={2026}
}

@inproceedings{xie2026llm,
  title={Llm-oriented token-adaptive knowledge distillation},
  author={Xie, Xurong and Xue, Zhucun and Wu, Jiafu and Li, Jian and Wang, Yabiao and Hu, Xiaobin and Liu, Yong and Zhang, Jiangning},
  booktitle={Proceedings of the AAAI Conference on Artificial Intelligence},
  volume={40},
  number={40},
  pages={34070--34078},
  year={2026}
}

@inproceedings{li2025bild,
  title={Bild: Bi-directional logits difference loss for large language model distillation},
  author={Li, Minchong and Zhou, Feng and Song, Xiaohui},
  booktitle={Proceedings of the 31st International Conference on Computational Linguistics},
  pages={1168--1182},
  year={2025}
}

@inproceedings{kim2016sequence,
  title={Sequence-level knowledge distillation},
  author={Kim, Yoon and Rush, Alexander M},
  booktitle={Proceedings of the 2016 conference on empirical methods in natural language processing},
  pages={1317--1327},
  year={2016}
}

@inproceedings{gu2024miniplm,
  title={{MiniPLM}: Knowledge Distillation for Pre-training Language Models},
  author={Gu, Yuxian and Zhou, Hao and Meng, Fandong and Zhou, Jie and Huang, Minlie},
  booktitle={The Thirteenth International Conference on Learning Representations},
  year={2025}
}

@inproceedings{gu2024minillm,
  title={Minillm: Knowledge distillation of large language models},
  author={Gu, Yuxian and Dong, Li and Wei, Furu and Huang, Minlie},
  booktitle={International Conference on Learning Representations},
  volume={2024},
  pages={32694--32717},
  year={2024}
}

@inproceedings{
agarwal2024onpolicy,
title={On-Policy Distillation of Language Models: Learning from Self-Generated Mistakes},
author={Rishabh Agarwal and Nino Vieillard and Yongchao Zhou and Piotr Stanczyk and Sabela Ramos Garea and Matthieu Geist and Olivier Bachem},
booktitle={The Twelfth International Conference on Learning Representations},
year={2024},
url={https://openreview.net/forum?id=3zKtaqxLhW}
}

@inproceedings{wen2023f,
  title={F-divergence minimization for sequence-level knowledge distillation},
  author={Wen, Yuqiao and Li, Zichao and Du, Wenyu and Mou, Lili},
  booktitle={Proceedings of the 61st Annual Meeting of the Association for Computational Linguistics (Volume 1: Long Papers)},
  pages={10817--10834},
  year={2023}
}

@inproceedings{menon2021statistical,
  title={A Statistical Perspective on Distillation},
  author={Menon, Aditya K and Rawat, Ankit Singh and Reddi, Sashank and Kim, Seungyeon and Kumar, Sanjiv},
  booktitle={Proceedings of the 38th International Conference on Machine Learning},
  volume={139},
  series={Proceedings of Machine Learning Research},
  pages={7632--7642},
  year={2021},
  publisher={PMLR}
}

@inproceedings{mirzadeh2020improved,
  title={Improved knowledge distillation via teacher assistant},
  author={Mirzadeh, Seyed Iman and Farajtabar, Mehrdad and Li, Ang and Levine, Nir and Matsukawa, Akihiro and Ghasemzadeh, Hassan},
  booktitle={Proceedings of the AAAI conference on artificial intelligence},
  volume={34},
  number={04},
  pages={5191--5198},
  year={2020}
}

@inproceedings{zhang2025towards,
  title={Towards the law of capacity gap in distilling language models},
  author={Zhang, Chen and Li, Qiuchi and Song, Dawei and Ye, Zheyu and Gao, Yan and Hu, Yao},
  booktitle={Proceedings of the 63rd Annual Meeting of the Association for Computational Linguistics (Volume 1: Long Papers)},
  pages={22504--22528},
  year={2025}
}

@article{stanton2021does,
  title={Does knowledge distillation really work?},
  author={Stanton, Samuel and Izmailov, Pavel and Kirichenko, Polina and Alemi, Alexander A and Wilson, Andrew G},
  journal={Advances in neural information processing systems},
  volume={34},
  pages={6906--6919},
  year={2021}
}

@inproceedings{zhao2022decoupled,
  title={Decoupled knowledge distillation},
  author={Zhao, Borui and Cui, Quan and Song, Renjie and Qiu, Yiyu and Liang, Jiajun},
  booktitle={2022 IEEE/CVF Conference on Computer Vision and Pattern Recognition (CVPR)},
  pages={11943--11952},
  year={2022},
  organization={IEEE}
}

@article{gao2020pile,
  title={The pile: An 800gb dataset of diverse text for language modeling},
  author={Gao, Leo and Biderman, Stella and Black, Sid and Golding, Laurence and Hoppe, Travis and Foster, Charles and Phang, Jason and He, Horace and Thite, Anish and Nabeshima, Noa and others},
  journal={arXiv preprint arXiv:2101.00027},
  year={2020}
}

@article{biderman2022datasheet,
  title={Datasheet for the pile},
  author={Biderman, Stella and Bicheno, Kieran and Gao, Leo},
  journal={arXiv preprint arXiv:2201.07311},
  year={2022}
}

@article{llama2023llama,
  title={{LLaMA}: Open and Efficient Foundation Language Models},
  author={Touvron, Hugo and Lavril, Thibaut and Izacard, Gautier and Martinet, Xavier and Lachaux, Marie-Anne and Lacroix, Timoth{\'e}e and Rozi{\`e}re, Baptiste and Goyal, Naman and Hambro, Eric and Azhar, Faisal and Rodriguez, Aurelien and Joulin, Armand and Grave, Edouard and Lample, Guillaume},
  journal={arXiv preprint arXiv:2302.13971},
  year={2023}
}

@inproceedings{groeneveld2024olmo,
  title={OLMo: Accelerating the science of language models},
  author={Groeneveld, Dirk and Beltagy, Iz and Walsh, Evan and Bhagia, Akshita and Kinney, Rodney and Tafjord, Oyvind and Jha, Ananya and Ivison, Hamish and Magnusson, Ian and Wang, Yizhong and others},
  booktitle={Proceedings of the 62nd annual meeting of the association for computational linguistics (volume 1: long papers)},
  pages={15789--15809},
  year={2024}
}

@article{gao2021framework,
  title={A framework for few-shot language model evaluation},
  author={Gao, Leo and Tow, Jonathan and Biderman, Stella and Black, Sid and DiPofi, Anthony and Foster, Charles and Golding, Laurence and Hsu, Jeffrey and McDonell, Kyle and Muennighoff, Niklas and others},
  journal={Zenodo},
  year={2021}
}

@inproceedings{zellers2019hellaswag,
  title={Hellaswag: Can a machine really finish your sentence?},
  author={Zellers, Rowan and Holtzman, Ari and Bisk, Yonatan and Farhadi, Ali and Choi, Yejin},
  booktitle={Proceedings of the 57th annual meeting of the association for computational linguistics},
  pages={4791--4800},
  year={2019}
}

@inproceedings{paperno2016lambada,
  title={The LAMBADA dataset: Word prediction requiring a broad discourse context},
  author={Paperno, Denis and Kruszewski, Germ{\'a}n and Lazaridou, Angeliki and Pham, Ngoc-Quan and Bernardi, Raffaella and Pezzelle, Sandro and Baroni, Marco and Boleda, Gemma and Fern{\'a}ndez, Raquel},
  booktitle={Proceedings of the 54th annual meeting of the association for computational linguistics (volume 1: Long papers)},
  pages={1525--1534},
  year={2016}
}

@article{sakaguchi2021winogrande,
  title={Winogrande: An adversarial winograd schema challenge at scale},
  author={Sakaguchi, Keisuke and Bras, Ronan Le and Bhagavatula, Chandra and Choi, Yejin},
  journal={Communications of the ACM},
  volume={64},
  number={9},
  pages={99--106},
  year={2021},
  publisher={ACM New York, NY, USA}
}

@inproceedings{mihaylov2018can,
  title={Can a suit of armor conduct electricity? a new dataset for open book question answering},
  author={Mihaylov, Todor and Clark, Peter and Khot, Tushar and Sabharwal, Ashish},
  booktitle={Proceedings of the 2018 conference on empirical methods in natural language processing},
  pages={2381--2391},
  year={2018}
}

@article{clark2018think,
  title={Think you have solved question answering? try arc, the ai2 reasoning challenge},
  author={Clark, Peter and Cowhey, Isaac and Etzioni, Oren and Khot, Tushar and Sabharwal, Ashish and Schoenick, Carissa and Tafjord, Oyvind},
  journal={arXiv preprint arXiv:1803.05457},
  year={2018}
}

@inproceedings{bisk2020piqa,
  title={Piqa: Reasoning about physical commonsense in natural language},
  author={Bisk, Yonatan and Zellers, Rowan and Gao, Jianfeng and Choi, Yejin and others},
  booktitle={Proceedings of the AAAI conference on artificial intelligence},
  volume={34},
  number={05},
  pages={7432--7439},
  year={2020}
}

@inproceedings{sap2019social,
  title={Social IQa: Commonsense reasoning about social interactions},
  author={Sap, Maarten and Rashkin, Hannah and Chen, Derek and Le Bras, Ronan and Choi, Yejin},
  booktitle={Proceedings of the 2019 conference on empirical methods in natural language processing and the 9th international joint conference on natural language processing (EMNLP-IJCNLP)},
  pages={4463--4473},
  year={2019}
}

@inproceedings{mostafazadeh2016corpus,
  title={A corpus and cloze evaluation for deeper understanding of commonsense stories},
  author={Mostafazadeh, Nasrin and Chambers, Nathanael and He, Xiaodong and Parikh, Devi and Batra, Dhruv and Vanderwende, Lucy and Kohli, Pushmeet and Allen, James},
  booktitle={Proceedings of the 2016 Conference of the North American Chapter of the Association for Computational Linguistics: Human Language Technologies},
  pages={839--849},
  year={2016}
}

@inproceedings{loshchilov2017decoupled,
  title={Decoupled Weight Decay Regularization},
  author={Loshchilov, Ilya and Hutter, Frank},
  booktitle={International Conference on Learning Representations},
  year={2019}
}

\clearpage

\appendix

\numberwithin{equation}{section}
\counterwithin{table}{section}
\counterwithin{figure}{section}

\section{Derivation of the Local--Rest KL Decomposition}
\label{app:kl_decomposition}

For compactness, we omit the prediction-position index \(u\). Let \(P^T\) and \(P^S\) be the teacher and student full-vocabulary distributions, with components \(p_i^T\) and \(p_i^S\). Given a partition of the vocabulary into a local token set \(\I\) and its complement \(\R=\V\setminus\I\), define
\[
    \alpha^T=\sum_{i\in\I}p_i^T,
    \quad
    \bar{\alpha}^T=1-\alpha^T,
    \quad
    \alpha^S=\sum_{i\in\I}p_i^S,
    \quad
    \bar{\alpha}^S=1-\alpha^S.
\]
For each token \(i\in\I\), define the teacher and student conditional probabilities as
\[
    \tilde{p}_{i}^{T,\I}=\frac{p_i^T}{\alpha^T},
    \qquad
    \tilde{p}_{i}^{S,\I}=\frac{p_i^S}{\alpha^S}.
\]
These probabilities define the conditional distributions
\(\tilde{P}^{T,\I}\) and \(\tilde{P}^{S,\I}\) over \(\I\). For each token
\(i\in\R\), define the corresponding conditional probabilities as
\[
    \tilde{p}_{i}^{T,\R}=\frac{p_i^T}{\bar{\alpha}^T},
    \qquad
    \tilde{p}_{i}^{S,\R}=\frac{p_i^S}{\bar{\alpha}^S}.
\]
These probabilities define the conditional distributions \(\tilde{P}^{T,\R}\) and \(\tilde{P}^{S,\R}\) over \(\R\). Starting from the full-vocabulary forward KL,
\begin{align}
    \KL(P^T\|P^S)
    &=
    \sum_{i\in\I}p_i^T\log\frac{p_i^T}{p_i^S}
    +
    \sum_{i\in\R}p_i^T\log\frac{p_i^T}{p_i^S}
    \nonumber\\
    &=
    \alpha^T
    \sum_{i\in\I}
    \tilde{p}_{i}^{T,\I}
    \log
    \frac{\alpha^T\tilde{p}_{i}^{T,\I}}
         {\alpha^S\tilde{p}_{i}^{S,\I}}
    +
    \bar{\alpha}^T
    \sum_{i\in\R}
    \tilde{p}_{i}^{T,\R}
    \log
    \frac{\bar{\alpha}^T\tilde{p}_{i}^{T,\R}}
         {\bar{\alpha}^S\tilde{p}_{i}^{S,\R}}
    \nonumber\\
    &=
    \alpha^T\log\frac{\alpha^T}{\alpha^S}
    +
    \alpha^T
    \KL(\tilde{P}^{T,\I}\|\tilde{P}^{S,\I})
    +
    \bar{\alpha}^T\log\frac{\bar{\alpha}^T}{\bar{\alpha}^S}
    +
    \bar{\alpha}^T
    \KL(\tilde{P}^{T,\R}\|\tilde{P}^{S,\R}).
    \label{eq:appendix_local_rest_decomposition}
\end{align}
Because
\[
    \KL(b^T\|b^S)
    =
    \alpha^T\log\frac{\alpha^T}{\alpha^S}
    +
    \bar{\alpha}^T\log\frac{\bar{\alpha}^T}{\bar{\alpha}^S},
\]
where \(b^T=(\alpha^T,\bar{\alpha}^T)\) and
\(b^S=(\alpha^S,\bar{\alpha}^S)\), we obtain
\begin{equation}
    \KL(P^T\|P^S)
    =
    \KL(b^T\|b^S)
    +
    \alpha^T\KL(\tilde{P}^{T,\I}\|\tilde{P}^{S,\I})
    +
    \bar{\alpha}^T\KL(\tilde{P}^{T,\R}\|\tilde{P}^{S,\R}).
    \label{eq:appendix_final_decomposition}
\end{equation}
Equation~\eqref{eq:appendix_final_decomposition} is the exact local--rest decomposition of the full-vocabulary forward KL. In the ALD objective defined in Eq.~\eqref{eq:ald}, the mass term is retained, whereas the coefficients \(\alpha^T\) and \(\bar{\alpha}^T\) are removed from the local- and rest-conditional divergences, respectively. Therefore, the exact decomposition provides the structural basis for ALD, but ALD defines a distinct objective because the teacher-mass coefficients of the local- and rest-conditional terms are replaced by unit coefficients.

\FloatBarrier

\section{Model Configurations}
\label{app:model_configurations}

The teacher and both students follow the same Qwen-family decoder-only architecture and use the same tokenizer and vocabulary of 151{,}936 tokens. The pretrained teacher remains fixed throughout every distillation run. The public student configurations are used only to determine the model architectures; the released student weights are not used, and both students are initialized randomly before pre-training. To control for variation due to random initialization, we use a fixed random seed (seed \(= 1234\)) for model initialization. For each student architecture, all compared methods therefore start from identical initial parameter values. In particular, all 200M experiments share the same initial 200M student parameters, and all 500M experiments share the same initial 500M student parameters. All three configurations support context lengths greater than the 512-token model-input length used in our experiments, so the model-supported context length should not be conflated with the experimental sequence length.

Table~\ref{tab:model_configurations} lists the architecture values used in the reported experiments. The 500M student preserves the teacher depth and attention-head configuration while reducing the hidden and feed-forward dimensions. The 200M student reduces both depth and width. Consequently, the two settings create different capacity gaps while keeping the architecture family, tokenizer, and output space fixed.

\begin{center}
\captionof{table}{Architectural configurations of the teacher and student models used in the pre-training distillation experiments. The student architectures follow the public Qwen-based configurations, but their released weights are not used; both students are initialized randomly before pre-training.}
\label{tab:model_configurations}
\begin{minipage}{\textwidth}
\centering
\small
\setlength{\tabcolsep}{5pt}
\renewcommand{\arraystretch}{1.10}
\begin{adjustbox}{max width=\linewidth}
\begin{tabular}{@{}llrrrrrrrl@{}}
\toprule
\textbf{Model} & \textbf{Role} & \textbf{Hidden} & \textbf{FFN} &
\textbf{Layers} & \textbf{Heads} & \textbf{KV heads} &
\textbf{Vocabulary} & \textbf{Parameters} & \textbf{Tied emb.} \\
\midrule
Qwen1.5-1.8B & Teacher & 2048 & 5504 & 24 & 16 & 16 & 151{,}936 & 1.837B & No \\
Pretrain-Qwen-500M & Student & 1024 & 2816 & 24 & 16 & 16 & 151{,}936 & 464.0M & Yes \\
Pretrain-Qwen-200M & Student & 768 & 2112 & 12 & 12 & 12 & 151{,}936 & 203.4M & Yes \\
\bottomrule
\end{tabular}
\end{adjustbox}
\end{minipage}
\end{center}

\FloatBarrier

\FloatBarrier

\section{Training Data Construction}
\label{app:training_data}

\subsection{Source Corpus and Document Selection}

The pre-training text is drawn from Pile
Uncopyrighted\footnote{\url{https://huggingface.co/datasets/monology/pile-uncopyrighted}},
a filtered version of The Pile \citep{gao2020pile,biderman2022datasheet}
obtained by removing the Books3, BookCorpus2, OpenSubtitles, YTSubtitles,
and OWT2 subsets. We stream the dataset with the Hugging Face \texttt{datasets} library without upstream shuffling and retain the first 2{,}000{,}000 records in the resulting stream order. These records correspond to approximately 2.8B tokenizer tokens before document-boundary EOS insertion and final training-budget truncation. From each record, only the \texttt{text} field is used; no metadata field is used for tokenization, example construction, or training.

\subsection{Tokenization and Boundary-Aware Chunk Construction}

Each document's \texttt{text} field is encoded with the shared Qwen tokenizer used by the teacher and both students, with automatic special-token insertion disabled, and a single EOS token is appended to mark the document boundary. Token IDs are accumulated in a buffer, and each stored example contains at most 513 token IDs. For a stored sequence \(s=(s_1,\ldots,s_L)\), \(L\leq513\), the model input is \(s_{1:L-1}\) and the corresponding next-token targets are \(s_{2:L}\). Hence, a full-length stored example provides 512 model-input positions and 512 next-token targets, matching the maximum model-input length used in training.

When a candidate chunk reaches the maximum stored length, the procedure searches backward within the chunk for the latest acceptable boundary, considering an EOS boundary, a newline boundary, or a sentence-ending candidate that passes checks against abbreviations and mid-word tokenization. Tokens following the selected boundary are returned to the buffer and become the prefix of the next example. This boundary-aware construction reduces unnatural truncation while keeping examples close to the maximum length; it does not change the one-token next-token alignment or the definition of valid positions in Eq.~\eqref{eq:valid_token_positions}.

\subsection{Data Ordering and Token Budgets}

The preprocessing logs record the number and lengths of the constructed sequences together with any padding introduced when shorter examples are batched. The full-budget processed training stream is constructed once, and the shorter controlled runs use the corresponding prefix of the same stream rather than a separately generated corpus. Thus, the approximately 550M-token setting is an exact prefix of the approximately 1B-token setting. The training order is precomputed and shuffling is disabled, so all methods compared under the same budget observe the same sequence order, thereby removing data-order variation as a source of confounding.

With an effective batch size of 16 sequences and a maximum of 512 model-input tokens per sequence, 67.2K optimizer updates correspond to \(67{,}200\times16\times512=550{,}502{,}400\) nominal model-input tokens, which we report as the approximately 550M-token setting. Similarly, 126.4K updates correspond to \(126{,}400\times16\times512=1{,}035{,}468{,}800\) nominal model-input tokens, which we report as the approximately 1B-token setting. These counts are nominal: the number of positions that actually contribute to the objective can be smaller because padded or otherwise ignored target positions are excluded from \(\Tbatch\).

\FloatBarrier

\section{Evaluation Details}
\label{app:evaluation_details}

\subsection{Benchmark Tasks, Splits, and Metrics}

Table~\ref{tab:evaluation_suite} reports the LM Evaluation Harness task identifiers and the numbers of examples used in the executed protocol. The held-out column denotes the examples used to produce Table~\ref{tab:baseline_comparison}, and the validation column denotes the examples reserved for the controlled 550M-token analyses.

To guarantee a labeled held-out set for every task, we assign the held-out split by a fixed fallback rule: we first take the official test split, and only when that split is unavailable or its public labels are unusable do we fall back to the public validation split as the held-out set. For five of the nine tasks (LAMBADA-OpenAI, OpenBookQA, ARC-Easy, ARC-Challenge, and StoryCloze-2016) the official test split is usable and is used directly for held-out reporting. For the remaining four (HellaSwag, WinoGrande, PIQA, and SocialIQA) the official test labels are not usable, so the public validation split is promoted to the held-out set.

A separate validation set is reserved only for HellaSwag, ARC-Easy, and PIQA, which are the three tasks used in the 550M-token controlled analyses (the adaptive-budget study in Table~\ref{tab:validation_fixed_budget_ablation} and the anchoring study in Table~\ref{tab:validation_grounding_ablation}). The source of the separate validation set depends on which labeled split remains available after assigning the held-out set. For ARC-Easy the official test split is usable and serves as the held-out set (2{,}376 examples), so the official validation split remains free and is used for the controlled analyses (570 examples); its validation count is therefore smaller than its held-out count. For HellaSwag and PIQA the official test labels are not usable, so the public validation split is promoted to the held-out set (10{,}042 and 1{,}838 examples), and the larger labeled training split is then reserved for the controlled analyses (39{,}905 and 16{,}113 examples). For the remaining tasks, we do not reserve a separate validation set. In all cases the reserved validation examples are used only for the controlled mechanism analyses; they never update model parameters and therefore remain distinct from downstream fine-tuning. We report each split by the role it plays in our protocol rather than relabeling a public validation split as an official test set.

\begin{table}[!tbp]
\caption{Zero-shot evaluation suite, LM Evaluation Harness task identifiers, and the numbers of examples used for validation and held-out reporting. Held-out sets use the official test split where usable (LAMBADA-OpenAI, OpenBookQA, ARC-Easy, ARC-Challenge, StoryCloze-2016) and the public validation split otherwise (HellaSwag, WinoGrande, PIQA, SocialIQA). A dash in the validation column indicates that no separate validation set is reserved. Where a validation set is reserved, its source is the training split for HellaSwag and PIQA and the official validation split for ARC-Easy; these validation examples are used only for the controlled 550M-token analyses.}
\label{tab:evaluation_suite}
\centering
\scriptsize
\setlength{\tabcolsep}{3.8pt}
\renewcommand{\arraystretch}{1.15}
\begin{adjustbox}{max width=\linewidth}
\begin{tabular}{@{}>{\raggedright\arraybackslash}m{2.15cm}>{\raggedright\arraybackslash}m{2.25cm}>{\centering\arraybackslash}m{1.10cm}>{\centering\arraybackslash}m{1.10cm}>{\raggedright\arraybackslash}m{4.05cm}>{\raggedright\arraybackslash}m{2.35cm}@{}}
\toprule
\textbf{Benchmark} & \textbf{LM-Eval task ID} & \textbf{Validation} & \textbf{Held-out} & \textbf{Primary capability} & \textbf{Metric} \\
\midrule
HellaSwag & \texttt{hellaswag} & 39{,}905 & 10{,}042 & Commonsense continuation selection & Multiple-choice accuracy \\
LAMBADA-OpenAI & \texttt{lambada\_openai} & -- & 5{,}153 & Broad-context next-word prediction & Exact next-word accuracy \\
WinoGrande & \texttt{winogrande} & -- & 1{,}267 & Commonsense coreference reasoning & Multiple-choice accuracy \\
OpenBookQA & \texttt{openbookqa} & -- & 500 & Multiple-choice science question answering & Multiple-choice accuracy \\
ARC-Easy & \texttt{arc\_easy} & 570 & 2{,}376 & Grade-school science reasoning, easy set & Multiple-choice accuracy \\
ARC-Challenge & \texttt{arc\_challenge} & -- & 1{,}172 & Grade-school science reasoning, challenge set & Multiple-choice accuracy \\
PIQA & \texttt{piqa} & 16{,}113 & 1{,}838 & Physical commonsense reasoning & Multiple-choice accuracy \\
SocialIQA & \texttt{social\_iqa} & -- & 1{,}954 & Social commonsense reasoning & Multiple-choice accuracy \\
StoryCloze-2016 & \texttt{storycloze\_2016} & -- & 1{,}000 & Narrative ending selection & Multiple-choice accuracy \\
\bottomrule
\end{tabular}
\end{adjustbox}
\end{table}
\FloatBarrier

\begin{figure}
    \centering
    \includegraphics[width=0.90\textwidth, keepaspectratio]{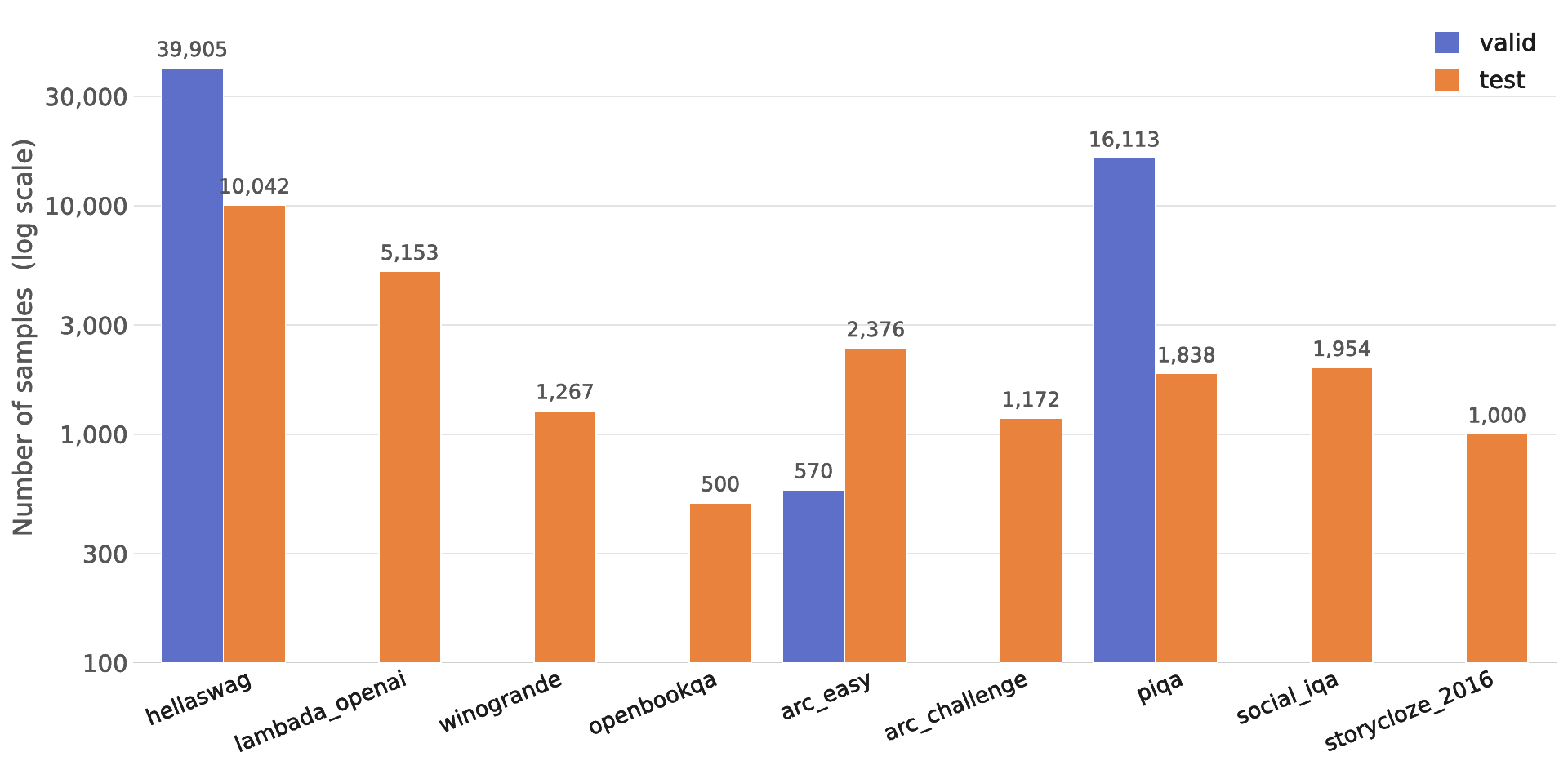}
    \caption{
    Validation and held-out evaluation set sizes for the nine downstream benchmarks, on a logarithmic vertical axis. Blue bars denote the validation sets reserved for the controlled 550M-token studies; orange bars denote the held-out sets used in the nine-benchmark comparison.
    }
    \label{fig:evaluation_sizes}
\end{figure}

Figure~\ref{fig:evaluation_sizes} visualizes the counts in Table~\ref{tab:evaluation_suite} on a logarithmic axis, used because the sets span from 39{,}905 examples down to 500. Blue bars appear only for HellaSwag, ARC-Easy, and PIQA, matching the validation sets reserved for the two 550M-token controlled studies; orange bars denote the held-out sets used in the nine-task comparison. The plot is descriptive rather than a weighting rule: because ``Avg.'' gives every benchmark equal weight, HellaSwag and PIQA do not dominate the aggregate merely because they contain more examples.

\FloatBarrier

\subsection{Controlled Mechanism Analysis}

Tables~\ref{tab:validation_fixed_budget_ablation} and~\ref{tab:validation_grounding_ablation} use the same ARC-Easy, HellaSwag, and PIQA validation sets after approximately 550M training tokens (67.2K optimizer updates), with the arithmetic mean computed over these three task scores at equal task weight. The fixed-budget sweep, the candidate-set anchoring variants, and the full ALRA row are therefore compared at the same checkpoint and on the same validation examples. These are mechanism-oriented controlled analyses; they are kept separate from the nine-task held-out results in Table~\ref{tab:baseline_comparison}. Table~\ref{tab:pair_weighting_ablation}, in contrast, is evaluated after the approximately 1B-token run and reports the same nine-task held-out average as the main comparison.

\subsection{Zero-Shot Scoring and Averaging}

No labeled downstream example is used to update student parameters; each task is scored with its default LM Evaluation Harness configuration. We report multiple-choice accuracy for the eight selection tasks and exact next-word accuracy for LAMBADA-OpenAI. The final average gives each benchmark weight \(1/9\), irrespective of the number of examples in Table~\ref{tab:evaluation_suite}, so no task dominates the aggregate merely by being larger. The evaluation uses the latest LM Evaluation Harness commit available when the evaluations were run. To support reproducibility, we release the corresponding commit identifier and exported task configurations in our evaluation code.

\FloatBarrier

\section{Training and Implementation Details}
\label{app:implementation_details}

\subsection{Optimization Configuration}
\label{app:optimization}

All runs use the shared optimization configuration in Table~\ref{tab:optimization_configuration}. Students are optimized with AdamW~\citep{loshchilov2017decoupled} using \(\beta_1=0.9\), \(\beta_2=0.98\), \(\epsilon_{\mathrm{AdamW}}=10^{-6}\), and weight decay \(10^{-2}\). A micro-batch contains one sequence, and 16 gradient-accumulation steps give an effective batch of 16 sequences. Gradients are clipped to a maximum norm of 0.5. The learning rate is warmed up linearly for 512 updates to \(6\times10^{-4}\) and then follows a cosine schedule defined over the full 126.4K-update budget, with a terminal learning rate of \(6\times10^{-5}\). The controlled 67.2K-update runs terminate at the corresponding intermediate point of this same schedule. The teacher is evaluated without parameter updates. All methods compared under one setting use the same optimizer, sequence order, sequence length, and update budget.

\begin{table}[!tbp]
\caption{Shared optimization and sequence configuration used in the reported
experiments.}
\label{tab:optimization_configuration}
\centering
\small
\setlength{\tabcolsep}{8pt}
\renewcommand{\arraystretch}{1.10}
\begin{tabular}{@{}ll@{}}
\toprule
\textbf{Configuration item} & \textbf{Value} \\
\midrule
Optimizer & AdamW \\
AdamW coefficients & \(\beta_1=0.9,\ \beta_2=0.98\) \\
AdamW epsilon & \(10^{-6}\) \\
Weight decay & \(10^{-2}\) \\
Maximum model-input length & 512 tokens \\
Stored pre-shift sequence length & Up to 513 token IDs \\
Micro-batch size & 1 sequence \\
Gradient accumulation & 16 steps \\
Effective batch size & 16 sequences \\
Gradient clipping & 0.5 \\
Warmup & 512 updates \\
Peak learning rate & \(6\times10^{-4}\) \\
Final learning rate & \(6\times10^{-5}\) \\
Learning-rate decay & Cosine \\
Controlled-analysis checkpoint & 67.2K updates (approximately 550M tokens) \\
Full-run endpoint & approximately 126.4K updates (approximately 1B tokens) \\
\bottomrule
\end{tabular}
\end{table}
\FloatBarrier

\paragraph{Checkpoint reporting.}
All reported results are obtained from the checkpoint at the end of the assigned training-token budget: 67.2K optimizer updates for the controlled 550M-token analyses and 126.4K updates for the approximately 1B-token experiments. We do not select checkpoints based on downstream validation or held-out performance. The main comparison, the pairwise weighting study, and the controlled analyses use the split roles described in Appendix~\ref{app:evaluation_details}. Because the reported runs do not include across-seed statistics, we do not claim statistical significance for small numerical differences.

\subsection{ALRA Hyperparameters}
\label{app:alra_hyperparameters}

Table~\ref{tab:alra_hyperparameters} lists the ALRA hyperparameters and fixed-budget values used in the reported experiments and controlled analyses. ALRA uses \(d_{\min}=3\), \(d_{\max}=25\), and the fixed-budget sweep \(d\in\{3,7,9,15,19,25\}\). The probability-gap sensitivity in Eq.~\eqref{eq:unnormalized_pair_score} is \(\gamma=5\). The distillation and pairwise temperatures are both set to \(1.0\), the pairwise-loss coefficient to \(\lambda_{\mathrm{pair}}=1.0\), and the numerical-stability constant to \(\epsilon=10^{-6}\). The reported ALRA comparison does not add the optional causal language modeling term, so \(\lambda_{\mathrm{CE}}=0\).

\begin{table}[!tbp]
\caption{ALRA hyperparameters used in the reported experiments.}
\label{tab:alra_hyperparameters}
\centering
\small
\setlength{\tabcolsep}{8pt}
\renewcommand{\arraystretch}{1.10}
\begin{tabular}{@{}lll@{}}
\toprule
\textbf{Hyperparameter} & \textbf{Symbol} & \textbf{Value} \\
\midrule
Minimum local-set size & \(d_{\min}\) & 3 \\
Maximum local-set size & \(d_{\max}\) & 25 \\
Fixed-budget sweep & \(d\) & \(\{3,7,9,15,19,25\}\) \\
Probability-gap sensitivity & \(\gamma\) & 5 \\
Distillation temperature & \(\tau\) & 1.0 \\
Pairwise temperature & \(\tau_p\) & 1.0 \\
Pairwise-loss coefficient & \(\lambda_{\mathrm{pair}}\) & 1.0 \\
Causal-LM coefficient & \(\lambda_{\mathrm{CE}}\) & 0 \\
Numerical-stability constant & \(\epsilon\) & \(10^{-6}\) \\
\bottomrule
\end{tabular}
\end{table}
\FloatBarrier

The adaptive budget is computed from the teacher distribution conditioned on the student-proposed, teacher-anchored candidate set and normalized by the batch-average effective support; it is not computed from the student's full-vocabulary entropy. Likewise, \(\gamma\) controls only the probability-gap factor in SWPRA and should not be read as the coefficient of the complete pairwise loss, which is \(\lambda_{\mathrm{pair}}\).

\subsection{Baseline Implementations and Adaptations}
\label{app:baseline_implementations}

All baseline students use the same architectures, initial parameter values, processed corpus, data order, sequence length, update budget, and shared optimization configuration as ALRA. The methods differ only in objective construction and method-specific settings. Table~\ref{tab:baseline_configuration_summary} records the configurations used in the reported runs, and the paragraphs following it state where an implementation changes the original method's training regime or output space. The distillation and language modeling losses are combined linearly; Table~\ref{tab:baseline_configuration_summary} reports the weight on each term as \(w_{\mathrm{KD}}/w_{\mathrm{CE}}\), and unless a method retains an explicit language modeling term its CE weight is zero (pure distillation).

\begin{table}[!tbp]
\caption{Each row gives the unnormalized weights on the distillation (KD) and language modeling (CE) terms as \(w_{\mathrm{KD}}/w_{\mathrm{CE}}\); for example, (1.0/0.0) is pure distillation, (0.5/0.5) is equal weighting, and (0.0/1.0) is pure language modeling. Shared data and optimizer settings are given in Table~\ref{tab:optimization_configuration}; adaptations are described below.}
\label{tab:baseline_configuration_summary}
\centering
\scriptsize
\setlength{\tabcolsep}{4.2pt}
\renewcommand{\arraystretch}{1.13}
\begin{adjustbox}{max width=\linewidth}
\begin{tabular}{@{}l>{\raggedright\arraybackslash}p{1.0cm}c>{\raggedright\arraybackslash}p{2.3cm}>{\raggedright\arraybackslash}p{7.1cm}@{}}
\toprule
\textbf{Method} & \textbf{KD/CE weight} & \textbf{Temperature} &
\textbf{Restricted support} & \textbf{Objective used in the comparison} \\
\midrule
Pre-train w/o KD & 0.0/1.0 & \NA & None &
Standard next-token causal language modeling loss. \\
Vanilla KD & 0.5/0.5 & 0.5 & Full vocabulary &
Forward teacher-to-student KL combined with causal language modeling at equal weight. \\
PD & 1.0/0.0 & 0.5 & Top-\(p=0.95\), then top-\(k=50\) &
Teacher probabilities are processed by top-\(p\)/top-\(k\), renormalized, and matched with forward KL; the reported run uses a fixed mixture rather than the paper's WSD mixing schedule. \\
ATKD & 1.0/0.0 & 1.0 & Target vs.\ non-target vocabulary &
Token-dependent teaching split by a teacher-uncertainty coefficient into a hard \(50\%\) and easy \(50\%\) split, both weighted equally (0.5/0.5). \\
RLD & 1.0/0.0 & 1.0 & Label-refined full vocabulary &
Teacher logits refined using ground-truth labels, applied independently at every valid prediction position. \\
LDRLD & 1.0/0.5 & 1.0 & Local top-15; rest retained &
Recursive decoupling and recombining of logits with adaptive decay weighting, adapted from fixed-class image classification to vocabulary-token outputs. \\
TAD & 0.5/0.5 & 1.0 & Top-\(K=10\) head plus tail & Mass-decoupled head KL (MKLD) plus normalized tail KL (NKLD), with the tail term up-weighted by 2.0. \\
BiLD & 1.0/0.0 & 3.0 & Top-\(k=8\) teacher and student sets &
Sum of teacher-led (t-LD) and student-led (s-LD) pairwise logit-difference terms. \\
ALRA & 1.0/0.0 & Table~\ref{tab:alra_hyperparameters} & Adaptive \(d_u\in[3,25]\) plus full rest &
ALD (mass, local, rest) plus SWPRA (\(\gamma=5\)), with teacher top-1 anchoring and no ground-truth anchoring. \\
\bottomrule
\end{tabular}
\end{adjustbox}
\end{table}
\FloatBarrier

\paragraph{Pre-train w/o KD and Vanilla KD.}
Pre-train w/o KD uses only the next-token causal language modeling objective. Vanilla KD uses the full teacher and student vocabulary distributions and a forward-KL distillation term \citep{hinton2015distilling}, with the KL and causal language modeling components weighted equally (\(0.5/0.5\)) at temperature 0.5. Vanilla KD therefore provides the direct full-vocabulary reference for assessing whether the local/rest and relational structure of ALRA adds value beyond global matching.

\paragraph{PD.}
PD treats pre-training distillation as a design space over teacher-logit processing, loss selection, data/model scale, and the mixture of language modeling and KD \citep{peng2025pre}. The reported adaptation temperature-scales the teacher logits (\(T=0.5\)), first applies top-\(p=0.95\) truncation, and then applies top-\(k=50\) truncation before computing the forward KL. The original study further explores a warmup--stable--decay schedule for the KD coefficient \(\alpha\), in which \(\alpha\) increases during the first 10\% of training, remains at its maximum value through the 89\% stable phase, and decreases during the final 1\%. Because the stable phase constitutes the majority of training, we approximate this schedule with a fixed \(\alpha=1.0\), corresponding to pure distillation, while retaining the shared optimization schedule used by all compared methods. The reported PD result should therefore be interpreted as an adaptation under our common pre-training protocol rather than a reproduction of the full WSD-\(\alpha\) configuration.

\paragraph{ATKD.}
ATKD decomposes token-level KD into target-oriented and diversity-oriented terms linked by a teacher-uncertainty coefficient, and varies the teaching applied to tokens of different learning difficulty \citep{zhong2024revisiting}. In the reported configuration, tokens are split by the uncertainty coefficient into a hard half and an easy half. We use equal weights (0.5/0.5) for the two subsets under our common training protocol; this differs from the weighting used in the original ATKD configuration. The two subsets differ in their loss terms: the hard half uses both target-oriented and diversity-oriented KD, whereas the easy half omits the target-oriented term. ATKD therefore provides an adaptive-teaching baseline, but it does not adapt the number of local vocabulary alternatives in the manner of Eq.~\eqref{eq:adaptive_local_budget}.

\paragraph{RLD.}
RLD uses ground-truth label information to dynamically refine the teacher's logits, removing misleading signal at positions where the teacher's prediction conflicts with the label while preserving the teacher's class correlations \citep{sun2025knowledge}. In our adaptation, this refinement is applied independently at every valid next-token prediction position, with vocabulary indices playing the role of output classes. We do not attribute cross-layer supervision or an adaptive local budget to RLD.

\paragraph{LDRLD.}
LDRLD was proposed for fixed-class image classification and transfers fine-grained inter-class relations over the top-$d$ student logits through a recursive decoupling and recombination of the logits, with an adaptive decay weighting that emphasizes the most critical category pairs \citep{xu2025local}. In that setting it uses a small local size ($d=7$); because a language model vocabulary ($\sim$151k classes) is orders of magnitude larger than an image-label space, we enlarge it to $d=15$. Vocabulary indices at one autoregressive position play the role of output classes, and the language modeling term is retained with weight 0.5. Its entry in Table~\ref{tab:baseline_comparison} is therefore an adapted language model result, not a direct reproduction of the original image-classification experiment.

\paragraph{TAD.}
TAD separates a teacher-preferred head (top-\(K\), \(K=10\)) from the remaining vocabulary tail \citep{dasguptadon}, matching the head via a mass-decoupled KL divergence (MKLD) and the tail via a normalized KL divergence (NKLD) up-weighted by a factor 2.0, at temperature 1.0. In its larger-model experiments, TAD additionally adds a cosine loss between the student's and teacher's hidden states, following DistilBERT \citep{sanh2019distilbert}; because our teacher and students have different hidden sizes, this cosine loss cannot be applied directly without an additional representation-alignment mechanism. We therefore omit the hidden-state term and retain the logit-level MKLD and NKLD terms together with the causal language modeling objective, using equal weights of \(0.5/0.5\) for the distillation and causal-LM losses. For consistency with our common initialization protocol, we do not adopt TAD's teacher-derived attention initialization; the TAD student starts from the same random initialization as the other compared methods.

\paragraph{BiLD.}
BiLD constructs teacher-led pairwise logit differences on teacher top-\(k\) indices and student-led differences on student top-\(k\) indices, then sums the corresponding alignment terms (t-LD and s-LD) \citep{li2025bild}. The reported comparison applies the defining loss directly to randomly initialized students with \(k=8\) and temperature 3.0. BiLD was originally designed for a three-stage post-training pipeline (SFT of the teacher, SFT of the student, then the BiLD stage over the two fine-tuned models); as we distill from scratch during pre-training with no labeled fine-tuning, we drop the two SFT stages and apply the BiLD loss directly. The result should therefore be read as an adaptation to from-scratch pre-training rather than a claim about BiLD in its original setting.

\subsection{Hardware, Runtime, and Pairwise Cost}
\label{app:hardware}

The reported runs use NVIDIA GeForce RTX~4090 GPUs with 24GB of memory. A run uses either one GPU with the teacher and student co-located, or two GPUs with the teacher and student placed on separate devices; the two-device arrangement reduces memory pressure and is not data parallelism. For the full 1B-token runs (approximately 126.4K updates), the measured end-to-end wall-clock time on a single RTX~4090 with the teacher and student co-located is approximately three days for the 200M student and four days for the 500M student. These figures are observations for the stated hardware and code path rather than hardware-independent complexity estimates.

SWPRA constructs \(\binom{d_u}{2}\) two-token relations at each valid prediction position, so with \(d_{\max}=25\) the maximum is \(\binom{25}{2}=300\) pairs. Once the adaptive local token set \(\I_u\) is formed, the additional SWPRA pairwise construction scales as \(O(d_u^2)\), rather than quadratically with the full vocabulary size \(|\V|\). It therefore introduces a bounded pairwise overhead on top of the full-vocabulary computations already required by the distillation objective.

We quantify this overhead on a single RTX~4090 with the teacher and student co-located, at an effective batch size of 16 sequences and a maximum sequence length of 512. Under otherwise identical conditions, the 500M student requires 2.75\,s per optimizer step with ALRA versus 2.23\,s with Vanilla KD, corresponding to an approximately 23\% increase, while peak memory usage increases from 13.0\,GB to 16.5\,GB. For the 200M student, peak memory usage increases from 12.0\,GB to 14.0\,GB. These measurements quantify the additional computational and memory overhead of ALRA under the stated single-GPU configuration.

\FloatBarrier

\end{document}